\documentclass{article}

\usepackage[english]{babel}

\usepackage[letterpaper,top=2cm,bottom=2cm,left=3cm,right=3cm,marginparwidth=1.75cm]{geometry}

\usepackage{amsmath}
\usepackage{mathtools}
\usepackage{graphicx}
\usepackage{amsfonts}
\usepackage[colorlinks=true, allcolors=blue]{hyperref}
\usepackage{float} 
\usepackage{comment}
\usepackage{multirow} 
\usepackage[table]{xcolor}
\usepackage{siunitx}
\usepackage{enumitem}
\usepackage{tikz}
\definecolor{lightred}{HTML}{FFE6E6}
\usepackage{authblk}

\title{Physics-Informed Gaussian Process Regression for the Constitutive Modeling of Concrete: A Data-Driven Improvement to Phenomenological Models}
\author[1]{Chenyang Li}
\author[2]{Himanshu Sharma}
\author[4]{Youcai Wu}
\author[4]{Joseph Magallanes}
\author[1,3]{K.T. Ramesh}
\author[1,2]{Michael D. Shields}
\affil[1]{Hopkins Extreme Materials Institute, Johns Hopkins University, Baltimore, MD}
\affil[2]{Dept.\ of Civil and Systems Engineering, Johns Hopkins University, Baltimore, MD}
\affil[3]{Dept.\ of Mechanical Engineering, Johns Hopkins University, Baltimore, MD}
\affil[4]{Karagozian and Case, Inc., Glendale, CA}

\begin{document}
\maketitle

\begin{abstract}
Understanding and modeling the constitutive behavior of concrete is crucial for civil and defense applications, yet widely used phenomenological models such as Karagozian \& Case concrete (KCC) model depend on empirically calibrated failure surfaces that lack flexibility in model form and associated uncertainty quantification. This work develops a physics-informed framework that retains the modular elastoplastic structure of KCC model while replacing its empirical failure surface with a constrained Gaussian Process Regression (GPR) surrogate that can be learned directly from experimentally accessible observables. Triaxial compression data under varying confinement levels are used for training, and the surrogate is then evaluated at confinement levels not included in the training set to assess its generalization capability. Results show that an unconstrained GPR interpolates well near training conditions but deteriorates and violates essential physical constraints under extrapolation, even when augmented with simulated data. In contrast, a physics-informed GPR that incorporates derivative-based constraints aligned with known material behavior yields markedly better accuracy and reliability, including at higher confinement levels beyond the training range. Probabilistic enforcement of these constraints also reduces predictive variance, producing tighter confidence intervals in data-scarce regimes. Overall, the proposed approach delivers a robust, uncertainty-aware surrogate that improves generalization and streamlines calibration without sacrificing the interpretability and numerical efficiency of the KCC model, offering a practical path toward an improved constitutive models for concrete.
\end{abstract}

\section{Introduction}

Modeling the mechanical behavior of concrete is a long-lasting challenge in computational mechanics, driven by its importance across civil infrastructure and defense applications. Concrete is a quasi-brittle, heterogeneous material whose response depends on a complex interplay of microstructural features and loading history. Capturing this path-dependent behavior, across different stress states and strain rates, requires constitutive models that balance fidelity, interpretability, and generalization.

A significant body of work has focused on phenomenological models that account for the progressive degradation of concrete due to microcracking and irreversible deformation through state variables without explicitly modeling the corresponding physical mechanisms of deformation \cite{leelavanichkul2009survey,park2022review}. These models are typically formulated within the continuum mechanics framework by coupling plasticity and the evolution of damage variables. Notable examples include the concrete damaged plasticity model implemented in \textsc{Abaqus}, based on the work of Lubliner et al.~\cite{lubliner1989plastic} and Lee et al.~\cite{lee1998plastic}; the Karagozian \& Case concrete (KCC) model~\cite{malvar1997plasticity,wu2015numerical} and the concrete plastic damage model proposed by Grassl et al.~\cite{grassl2006damage} and Grassl and Jir{\'a}sek ~\cite{grassl2013cdpm2}, both implemented in \textsc{LS-DYNA}.
Among these, the KCC model~\cite{malvar1997plasticity,wu2015numerical} has become widely adopted for both quasi-static and dynamic analyses because it is computationally efficient and captures essential concrete behaviors such as confinement strengthening, softening behavior, and shear dilation \cite{wu2012performance,lee2021improved}.
However, phenomenological models, such as the KCC model, inherit a few fundamental limitations. First, its key constitutive relations, such as the failure surface and strength envelopes, are defined by fixed empirical equations, which restrict the model’s flexibility in representing complex material behavior. Second, the model requires extensive manual calibration of numerous parameters. This calibration is typically performed in an ad hoc manner, meaning that different analysts may obtain noticeably different parameter sets for the same material. Such dependence on user judgment and empirical tuning limits reproducibility and reduces predictive robustness.

Recent advances in data-driven and machine learning (ML) methods have opened new pathways for constitutive modeling. Instead of prescribing fixed empirical forms, these approaches aim to learn governing relationships directly from experimental or high-fidelity simulation data, without assuming closed-form functional dependencies. 
To date, data-driven ML approaches have largely been implemented using neural networks to model the complex mechanical behavior of diverse materials \cite{fuhg2024review}.
These methods can directly utilize data without expert knowledge, capture complex relationships, and provide computationally efficient predictions once trained. However, many of these models operate as black boxes and require large datasets to generalize effectively. Furthermore, their lack of physical grounding inhibits interpretability and limits their applicability in predictive and extrapolative modeling. These limitations have motivated growing interest in physics-informed formulations, using methods like Physics Informed Neural Networks (PINNs)~\cite{raissi2019physics} to embed mechanistic constraints into ML models to enhance both interpretability and predictive robustness \cite{haghighat2023constitutive, liu2020generic}. In parallel, researchers have explored modular ML approaches, where only a specific component of a constitutive law, such as the yield function in plasticity, is replaced by a data driven surrogate \cite{vlassis2022component,fuhg2023modular}. This modular strategy enhances interpretability and reduces data requirements while retaining the overall structure of the original constitutive frameworks.

Building in this direction, this work develops a physics-informed constitutive modeling framework that combines the interpretability of classical constitutive theory with the flexibility of data driven ML methods. Rather than attempting to learn the entire strain-to-stress mapping from scratch, we adopt a modular approach: retaining the general elastoplastic structure of the phenomenological model while replacing only its empirical failure surface function with a physically constrained ML model. This strategy addresses one of the most calibration-sensitive components of the model while preserving its physical interpretability and numerically robust performance.

To achieve this, we employ Gaussian Process Regression (GPR)~\cite{williams2006gaussian}—a nonparametric and probabilistic ML method that has recently emerged as a promising tool for constitutive modeling \cite{upadhyay2024physics,fuhg2022local}. GPR offers two key advantages highly relevant to concrete characterization: (1) it performs well in data-scarce regimes; and (2) it offers principled uncertainty quantification, providing predictive confidence. These characteristics are critical for modeling concrete, a material known for its inherent variability and high cost of experimental characterization. Furthermore, the GPR framework allows the incorporation of constraints~\cite{swiler2020survey} that enforce consistency with basic physics~(e.g. thermodynamics~\cite{sharma2024learning}) and mechanism-driven material behavior, such as the strengthening effect of confinement for concrete, thereby improving both accuracy and extrapolation reliability.

A major contribution of this study is the direct use of experimental data to train the GPR model. This contrasts with previous work that relies primarily on simulated data or internal variables inferred from constitutive models \cite{lyu2023machine}, and ensures that our surrogate remains grounded in measurable physical observables. Synthetic data generated using the calibrated KCC model are further employed to examine how data availability influences model performance.

In summary, this study presents an uncertainty-aware modeling framework that integrates experimental data, GPR learning, and physics-informed constraints into a unified constitutive formulation for concrete. By embedding a GPR surrogate within the established KCC model, we demonstrate that ML can enhance generalization, robustness, and interpretability, without discarding the accumulated knowledge of decades of constitutive modeling. The proposed framework represents an improvement of concrete constitutive modeling in regimes where data are limited and uncertainty plays a significant role.

\section{A Physics-informed, Data-Driven Phenomenological Model}

Phenomenological constitutive models play a critical role in large-scale simulations of material and structural failure. They represent the essential compromise between detailed physics-based mechanistic models for lower-scale physical mechanisms (e.g.~void nucleation, growth, and coalescence in ductile metals or micro-crack formation, growth, and propagation in concrete) and the need for fast computations that can scale to large, geometrically complex structures under extreme loads. They aim to capture the essential macroscale response up to and including the onset of material failure without explicit physics-based modeling of lower-scale mechanisms. This is accomplished by integrating empirical mathematical models (surrogates) that approximate the essential behavior (e.g.~strain hardening, softening, and material fracture) into existing continuum mechanics deformation theories (i.e.~J2 plasticity). 

An important limitation of these phenomenological models is the often ad hoc empirical nature of the mathematical expressions that drive critical phenomena like yield surface expansion and contraction. These expressions are often derived from convenience and are not necessarily guaranteed to ensure that the material satisfies basic physical laws. Moreover, these expressions are often highly parameterized functional forms that are difficult to calibrate and interpret. In this section, we derive a physics-informed, data-driven enhancement to an existing phenomenological model that alleviates these challenges. The model is constrained to satisfy certain essential and interpretable physical principles. It follows a non-parametric GPR form that facilitates direct calibration from strain-stress data and quantifies uncertainty in material response.

We begin by introducing the theoretical foundations and modeling strategy adopted in this study. Section 2.1 briefly reviews the KCC model and identifies the empirical failure surface, $\Gamma$, as the component to be enhanced. 
Section 2.2 establishes a data-driven mapping from experimental observables and formulates the surrogate relation for the failure surface as  $\Gamma=f(\varepsilon_v,\varepsilon_s,p)$. 
Section 2.3 reviews the standard GPR
and Section 2.4 details the integration of constrained GPR for the surrogate failure surface. 


\subsection{Overview of the Karagozian \& Case concrete (KCC) model}

\begin{figure}[htbp]
  \centering
  \includegraphics[width=\linewidth]{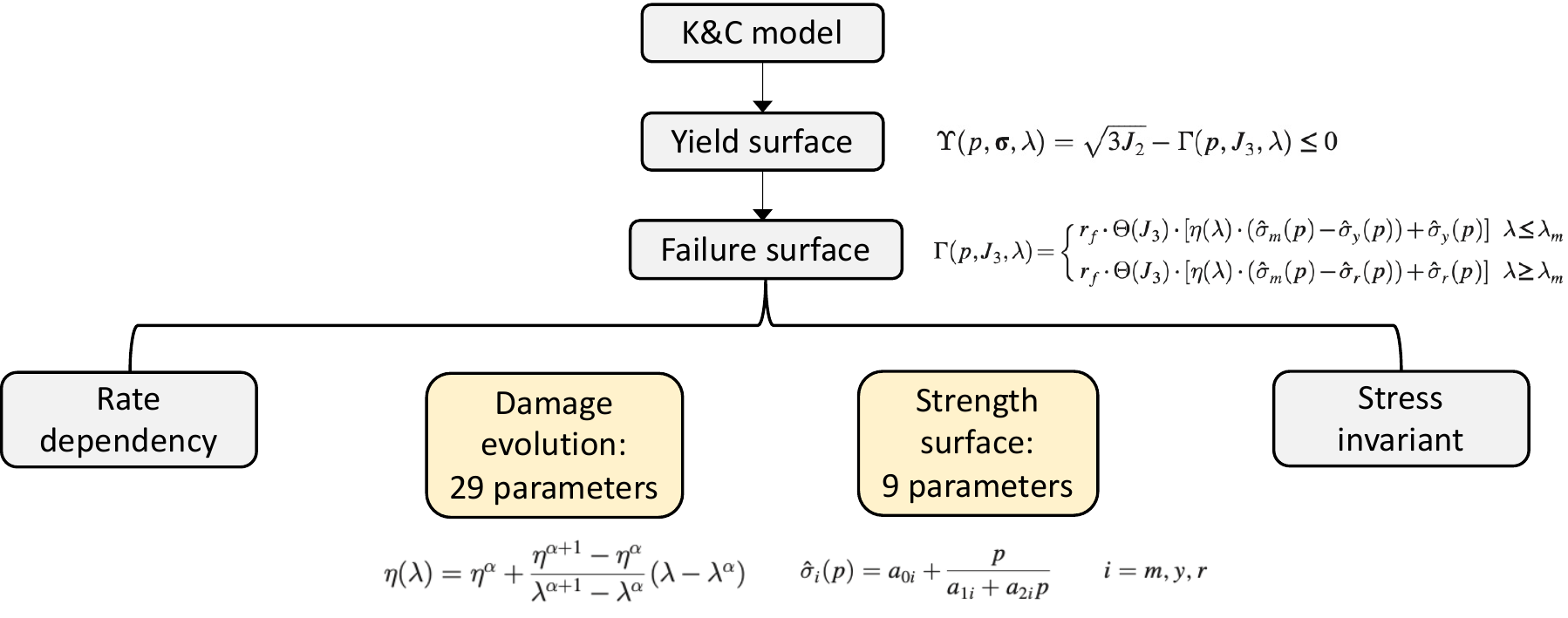}
  \caption{Schematic illustration of the Karagozian \& Case (KCC) model.}
  \label{fig:2_KC}
\end{figure}

The KCC model is a widely used phenomenological model for concrete and other quasi-brittle materials, which has been implemented in high-fidelity solvers such as LS-DYNA (\texttt{MAT072}) \cite{manual2014ls,wu2015numerical,wu2012performance} for broad industrial and research use. 
It is designed to capture the nonlinear, pressure- and rate-dependent responses of concrete over a broad range of loading conditions. 
The model adopts a plasticity framework in which plastic flow is governed by a pressure-dependent yield surface and a partially associated flow rule. The key components of the KCC model are illustrated in Figure \ref{fig:2_KC}.

The yield condition is defined by
\begin{equation}
\gamma(\boldsymbol{\sigma}, p, \lambda) = \sqrt{3J_2} - \Gamma(p, J_3, \lambda) \leq 0,
\label{KCC_Yield}
\end{equation}
where \( \boldsymbol{\sigma} \) is the Cauchy stress tensor, and \( J_2 = \frac{1}{2} \boldsymbol{\sigma}':\boldsymbol{\sigma}' \) and \( J_3 = \det(\boldsymbol{\sigma}') \) are the second and third invariants of the deviatoric stress tensor \( \boldsymbol{\sigma}' \). 
The pressure is defined as \( p = -\frac{1}{3}\mathrm{tr}(\boldsymbol{\sigma}) \), and \( \lambda \) is a scalar internal variable representing damage accumulated during plastic deformation.
This yield function can be decomposed into two distinct parts. The first, given by 
\( \sqrt{3J_2} \), is a stress norm rooted in classical plasticity, representing the physical deviatoric driving force. The second term, given by 
\( \Gamma(p, J_3, \lambda) \), is a phenomenological function defining a ``failure surface" which defines the material strength as a function of the state variables $p,\  J_3,\ \lambda$. The function \( \Gamma \) governs the evolution of the yield surface and is expressed as ~\cite{wu2015numerical}
\begin{equation}
\Gamma(p, J_3, \lambda) =
\begin{cases}
r_f \, \Theta(J_3) \,[\, \eta(\lambda)(\hat{\sigma}_m(p) - \hat{\sigma}_y(p)) + \hat{\sigma}_y(p) \,], & \lambda \leq \lambda_m, \\[6pt]
r_f \, \Theta(J_3) \,[\, \eta(\lambda)(\hat{\sigma}_m(p) - \hat{\sigma}_r(p)) + \hat{\sigma}_r(p) \,], & \lambda \geq \lambda_m,
\end{cases}
\end{equation}
where \( r_f \) is the dynamic increase factor, \( \Theta(J_3) \) controls the effect of the stress state through $J_3$, and \( \hat{\sigma}_y(p) \), \( \hat{\sigma}_m(p) \), and \( \hat{\sigma}_r(p) \) represent the pressure-dependent yield, maximum, and residual strength envelopes, respectively.
The function \( \Gamma(\cdot) \) combines several key phenomenological components:
\begin{itemize}[leftmargin=*]
  \item \textbf{Rate dependence:} A dynamic increase factor \( r_f \) accounts for strain-rate effects on strength.
  \item \textbf{Damage evolution:} A scalar damage variable \( \lambda \) serves as input to an interpolation function $\eta(\lambda)$ that interpolates between three pressure-dependent strength surfaces (discussed next). The evolution of \( \lambda \) is defined in terms of the rate of plastic strain tensor, and is given by
  \begin{equation}
      \dot{\lambda} = h(p)\dot{\bar{\varepsilon}}^p
  \end{equation}
  where $\dot{\bar{\varepsilon}}^p=\sqrt{2/3\dot{\boldsymbol{\varepsilon}}^p:\dot{\boldsymbol{\varepsilon}}^p}$ and 
  \begin{equation}
      h(p) = \begin{cases}
          [1+p/(r_ff_t)]^{-b_1}/r_f & p\ge 0\\
          [1+p/(r_ff_t)]^{-b_2}/r_f & p< 0
      \end{cases}
  \end{equation}
  where $b_1$ and $b_2$ are calibration parameters and $f_t$ is the concrete tensile strength.
  Plastic flow is governed by
  \begin{equation}
      \dot{\boldsymbol{\varepsilon}}^p = \dot{\mu} \dfrac{\partial \varphi(\boldsymbol{\sigma},p,\lambda)}{\partial \boldsymbol{\sigma}}
  \end{equation}
  where $\dot{\mu}$ is the plasticity consistency parameter and the plastic potential is given by
  \begin{equation}
      \varphi(\boldsymbol{\sigma}, p, \lambda) = \sqrt{3J_2} - \varpi\Gamma(p, J_3, \lambda)
  \end{equation}
  such that $\varpi$ controls the degree of associativity of the flow rule. In this work, a fixed $\varpi=0.5$ is used when applying the KCC model, although the original formulation allows for its dependence on the specific concrete type. 
  The interpolation function $\eta(\lambda)$ is defined by linear interpolation between points in a set of established $(\lambda, \eta)$ pairs. See \cite{wu2015numerical} for more details. In total, calibration of the damage evolution rule can require the estimation of up to 29 parameters. 
  \item \textbf{Strength surface functions:} The function $\eta(\lambda)$ controls the interpolation between three pressure dependent strength surfaces—yield (\( \hat{\sigma}_y(p) \)), maximum (\( \hat{\sigma}_m(p) \)), and residual (\( \hat{\sigma}_r(p) \)). When $\lambda\le \lambda_m$, a critical damage value, $\eta(\lambda)$ interpolates between the yield strength surface \( \hat{\sigma}_y(p) \) and the maximum strength surface \( \hat{\sigma}_m(p) \). When $\lambda>\lambda_m$, $\eta(\lambda)$ interpolates between the maximum strength surface \( \hat{\sigma}_m(p) \) and the residual strength surface \( \hat{\sigma}_r(p) \).  Each strength is defined by a pressure-dependent function with three user-defined parameters taking the form
  \begin{equation}
      \hat{\sigma}_i(p) = a_{0i}+\dfrac{p}{a_{1i}+a_{2i}p}; \qquad i = m, y, r
  \end{equation}
  leading to a total of nine parameters for the strength surfaces that must be calibrated. 
  \item \textbf{Stress-state dependence:} Dependence on the third invariant \( J_3 \) through the functional $\Theta(J_3)$ allows the model to distinguish between triaxial extension and compression, improving its ability to represent multiaxial stress conditions. The KCC model uses an expression developed by William and Warnke~\cite{william1975constitutive}, not repeated here for brevity. While $J_3$ dependence can, in principle, be included in the proposed model by including training data from different triaxial stress states, the present work considers only triaxial compression due to data availability. 
\end{itemize}

In total, the failure surface function \( \Gamma(\cdot) \) relies on all of the above fixed-form empirical relations and requires the calibration of more than 38 parameters. 
Such calibration demands substantial expert knowledge and is often performed in an ad hoc manner, meaning that different analysts may obtain significantly different parameter sets for the same material, depending on their experience and the available data. 
While the model is computationally efficient and versatile, its empirical function forms and user-dependent parameter calibrations limits its generalizability. 
Moreover, in data-scarce regimes, the model lacks a formal mechanism to quantify or propagate uncertainty, making it difficult to assess confidence in its predictions under untested loading conditions.

\subsection{Proposed physics-constrained, data-driven approach}

In the KCC yield condition (Eq.~\eqref{KCC_Yield}), the term \( \sqrt{3J_2} \) derives directly from classical plasticity theory and represents the physical deviatoric driving stress. 
In contrast, the function \( \Gamma(p,J_3,\lambda) \) is purely phenomenological, introduced to reproduce observed material responses through empirical fitting. 
Motivated by this distinction, we propose to replace the empirical failure surface function \( \Gamma(\cdot) \) with a data-driven, but physically constrained, surrogate. 
Such selective substitution of specific components within classical constitutive frameworks has proven effective in enhancing both interpretability and extensibility \cite{fuhg2022machine,bahmani2024discovering}. 

To construct the surrogate, we first define the input–output mapping. 
It is common to represent the constitutive response using the strain invariants \(\varepsilon_v\) (volumetric strain) and \(\varepsilon_s\) (deviatoric strain) \cite{vlassis2021sobolev,jung2006neural}. Since in the KCC model, $\Gamma$ is also pressure dependent, the surrogate seeks to model the mapping
\begin{equation}
    \Gamma = f(\varepsilon_v,\varepsilon_s,p).
    \label{eq:Gamma_mapping}
\end{equation}
Note that, more generally, $\Gamma(\cdot)$ may also be a function of the third invariant of the deviatoric stress tensor to distinguish between tensile and compressive states, but this is not considered here because we are interested primarily in compressive loading in this work. 

Although \(\Gamma\) is not directly measurable, it can be inferred from experimental observations under axisymmetric triaxial compression. 
For such tests,
\begin{equation}
    \sqrt{3J_2} = \sigma_a - \sigma_r = \sigma_q,
\end{equation}
where \(\sigma_a\) and \(\sigma_r\) are the axial and radial stresses, respectively, and \(\sigma_q\) denotes the deviatoric stress. 
Within the inelastic regime of the KCC model, the yield function (Eq.~\eqref{KCC_Yield} reduces to 
\begin{equation}
    \sqrt{3J_2} = \Gamma(p,J_3,\lambda),
\end{equation}
Accordingly, we obtain
\begin{equation}
    \sigma_q = \Gamma = f(\varepsilon_v,\varepsilon_s,p).
\end{equation}
By adopting the definitions of volumetric strain $\varepsilon_v=\varepsilon_a+2\varepsilon_r$ and deviatoric strain $\varepsilon_s=\varepsilon_a-\varepsilon_r$, where \(\varepsilon_a\) and \(\varepsilon_r\) denote the axial and radial strains measured in triaxial compression test, all quantities \((\varepsilon_v, \varepsilon_s, p, \sigma_q)\)
are directly experimental accessible. As a result, the surrogate mapping in Eq.~\eqref{eq:Gamma_mapping} can be trained entirely from experimental data, establishing the link between observable strain–stress responses and the evolution of the failure surface.

To improve performance and generalizability of the failure surface using a surrogate model, we establish the following criteria that such a model should satisfy:
\begin{itemize}[leftmargin=*]
  \item \textbf{Effective performance under data scarcity.} 
  The surrogate should be capable of accurate training and prediction in data-scarce regimes, where only limited experimental information is available. 
  Many existing studies rely on synthetic data from micromechanical or mesoscale simulations, allowing abundant training data but producing surrogates that represent the underlying model rather than the observed macroscale material behavior. 
  In contrast, this work expects to train the surrogate directly on experimental data to ensure consistency with physical observations. 
  However, available experimental datasets are often limited due to testing cost and measurement constraints. 
  Therefore, the ability to achieve reliable performance under data-scarce conditions is a primary objective of the proposed framework. In addition, the model should be sufficiently flexible to enable it to be enhanced by simulation data that may be available.

  \item \textbf{Quantification of model uncertainty.} 
  Concrete is a highly heterogeneous material whose macroscopic response is influenced by mix design, mixing procedure, and curing conditions \cite{price1951factors, lee2003prediction}. 
  Even under nominally identical conditions, significant scatter in the concrete mechanical properties can still occur. 
  Consequently, the surrogate must be able to quantify predictive uncertainty arising from parameter and model-form ambiguity arising from random variations in experimental observations and limited training data set size. 
  While parameter uncertainty has been widely studied and can be readily incorporated into parametric phenomenological models like the KCC model through Bayesian parameter estimation, the treatment of functional model-form uncertainty remains a key challenge \cite{sharma2024learning}. 
  The present framework aims to address both, integrating diverse data sources into a free-form, uncertainty-aware representation.

  \item \textbf{Compatibility with physical constraints.} 
  The surrogate should adhere to certain physics-based constraints to ensure generalizability and ensure realistic performance. 
  As the predictive capability of a surrogate model strongly depends on the amount and quality of training data, its performance can deteriorate rapidly when applied outside the training regime \cite{fuhg2023modular,fuhg2022physics,fuhg2023enhancing}. 
  On the other hand, several studies have demonstrated that incorporating physical constraints into classical machine learning models can significantly improve their generalizability and robustness \cite{liu2020generic,as2022mechanics,kalina2022automated,pierre2023principal,frankel2020tensor,fuhg2024stress}. 
  Therefore, the ability to embed physics-informed relations directly into the learning process is an essential feature of the proposed framework.

  \item \textbf{Mitigate ad hoc parameter calibration.} Calibration of the proposed model should follow a well-defined optimization process with a clearly defined objective function. This contrasts with large-scale ad hoc calibration processes routinely followed for phenomenological model calibration that are often driven by analyst experience and judgment.
\end{itemize}
We achieve these goals by expressing $\Gamma(\varepsilon_v, \varepsilon_s, p)$ through constrained Gaussian process regression as described in the following sections. 


\subsection{Review of Standard Gaussian Process Regression}
Gaussian Process Regression (GPR) is a nonparametric, Bayesian ML approach used to model unknown functions and their associated uncertainties. Unlike parametric regression models, GPR defines a distribution over functions and makes predictions by conditioning this distribution on observed data. This probabilistic framework makes GPR particularly appealing for constitutive modeling under uncertainty, especially in data-scarce regimes \cite{upadhyay2024physics}.

Consider a dataset with $N$ training inputs $\mathbf{X} = \{\mathbf{x}^{(1)}, \mathbf{x}^{(2)}, \dots, \mathbf{x}^{(N)}\}$, 
and corresponding outputs 
$\mathbf{y} = [y^{(1)}, y^{(2)}, \dots, y^{(N)}]^\top$. 
The goal of GPR is to learn a latent function $Y(\mathbf{x})$ that maps inputs to outputs, assuming that any finite collection of random variables 
$\{Y(\mathbf{x}^{(i)})\}$ follows a joint Gaussian distribution.

We define the Gaussian process prior as
\begin{equation}
Y(\mathbf{x}) \sim \mathcal{GP}\big[\,\mu(\mathbf{x}),\, K(\mathbf{x}, \mathbf{x}')\,\big],
\label{eq:GPR_prior}
\end{equation}
where $\mu(\mathbf{x})$ and $K(\mathbf{x}, \mathbf{x}')$ are the mean and covariance functions, respectively:
\begin{align}
\mu(\mathbf{x}) &= \mathbb{E}[Y(\mathbf{x})], \\
K(\mathbf{x}, \mathbf{x}') &= 
\mathbb{E}\big[\big(Y(\mathbf{x}) - \mu(\mathbf{x})\big)\big(Y(\mathbf{x}') - \mu(\mathbf{x}')\big)\big].
\end{align}

A widely used covariance function is the squared exponential (radial basis) kernel:
\begin{equation}
K(\mathbf{x}, \mathbf{x}') = 
\sigma_f^2 
\exp\!\left[-\frac{\|\mathbf{x}-\mathbf{x}'\|^2}{2l^2}\right]
\label{eq:GPR_kernel}
\end{equation}
where $l$ is the characteristic length scale and $\sigma_f^2$ is the signal variance.

For a new test point $\mathbf{x}_*$, the posterior predictive distribution for $Y(\mathbf{x}_*)$ conditioned on the training data $(\mathbf{X}, \mathbf{y})$ is given by
\begin{equation}
Y(\mathbf{x}_*) \,|\, \mathbf{y}, \mathbf{X} \sim 
\mathcal{N}\big[\, m(\mathbf{x}_*),\, s^2(\mathbf{x}_*) \,\big],
\label{eq:GPR_posterior}
\end{equation}
with predictive mean and variance:
\begin{align}
m(\mathbf{x}_*) &= 
\mu(\mathbf{x}_*) +
K(\mathbf{x}_*,\mathbf{X})
\big[K(\mathbf{X},\mathbf{X})+\sigma_n^2 I\big]^{-1}
\big[\mathbf{y}-\mu(\mathbf{X})\big], \label{eq:GPR_mean}\\[4pt]
s_c^2(\mathbf{x}_*) &=
K(\mathbf{x}_*,\mathbf{x}_*) -
K(\mathbf{x}_*,\mathbf{X})
\big[K(\mathbf{X},\mathbf{X})+\sigma_n^2 I\big]^{-1}
K(\mathbf{X},\mathbf{x}_*),
\label{eq:GPR_var}
\end{align}
with $\sigma_n^2$ denoting zero-mean Gaussian noise variance.
The set of hyperparameters, denoted by $\boldsymbol{\theta} = \{l, \sigma_f, \sigma_n\}$,
are estimated by maximizing the marginal likelihood, or equivalently minimizing its negative log form given by:
\begin{equation}
-\log p(\mathbf{y} \mid \mathbf{X}, \boldsymbol{\theta}) =
\frac{1}{2}(\mathbf{y}-\boldsymbol{\mu})^\top \mathbf{K}^{-1} (\mathbf{y}-\boldsymbol{\mu})
+\frac{1}{2}\log|\mathbf{K}|
+\frac{N}{2}\log(2\pi).
\label{eq:GPR_marglik}
\end{equation}

The standard GPR is unconstrained and in this work, we extend the standard GPR framework by incorporating
physics-informed constraints, to ensure the failure surface is consistent with certain established principles of mechanics as described next. 

\subsection{Physically-Constrained GPR-based Failure Surface}
\label{sec:PI-GPR}

The goal of the proposed GPR surrogate is to learn the nonlinear mapping
\begin{equation}
\Gamma = f(\varepsilon_v,\, \varepsilon_s,\, p),
\label{eq:gamma_relation}
\end{equation}
where $\varepsilon_v$ and $\varepsilon_s$ denote the volumetric and deviatoric strains, $p$ is the pressure, and $\Gamma$ represents the failure surface function in the KCC model. 
The surrogate aims to reproduce the physically consistent evolution of $\Gamma$ with respect to strain and confinement while quantifying predictive uncertainty.

\paragraph{GP prior and posterior.}
The prior distribution of $\Gamma$ over the input space $\mathbf{x}=(\varepsilon_v,\varepsilon_s,p)$ is defined as
\begin{equation}
\Gamma(\mathbf{x}) \sim \mathcal{GP}\!\big[\, \mu(\mathbf{x}),\, K(\mathbf{x},\mathbf{x}') \,\big],
\label{eq:PI_GPR_prior}
\end{equation}
where $\mu(\mathbf{x})$ is the mean function and $K(\mathbf{x},\mathbf{x}')$ is the covariance kernel, following Eq.~\eqref{eq:GPR_prior}.  
Given $N$ training samples $(\mathbf{X},\boldsymbol{\Gamma})$, the posterior predictive distribution at a new input $\mathbf{x}_*$ is
\begin{equation}
\Gamma(\mathbf{x}_*) \,|\, \boldsymbol{\Gamma},\mathbf{X}
\sim
\mathcal{N}\!\left[\, m_c(\mathbf{x}_*),\, s_c^2(\mathbf{x}_*) \,\right],
\label{eq:PI_GPR_posterior}
\end{equation}
where the posterior mean and variance are
\begin{align}
m_c(\mathbf{x}_*) &= 
\mu(\mathbf{x}_*) +
K(\mathbf{x}_*,\mathbf{X})
\big[K(\mathbf{X},\mathbf{X})+\sigma_n^2 I\big]^{-1}
\big[\boldsymbol{\Gamma}-\mu(\mathbf{X})\big], \label{eq:post_mean}\\[4pt]
s_c^2(\mathbf{x}_*) &=
K(\mathbf{x}_*,\mathbf{x}_*) -
K(\mathbf{x}_*,\mathbf{X})
\big[K(\mathbf{X},\mathbf{X})+\sigma_n^2 I\big]^{-1}
K(\mathbf{X},\mathbf{x}_*),
\label{eq:post_var}
\end{align}
again following directly from Eqs.~\eqref{eq:GPR_mean} and \eqref{eq:GPR_var}.

\paragraph{Kernel function.}
A stationary anisotropic squared-exponential kernel is adopted taking the form:
\begin{equation}
K(\mathbf{x},\mathbf{x}')=
\sigma_f^2
\exp\!\left[-\sum_{j=1}^{3}\frac{(x_j-x'_j)^2}{2\ell_j^2}\right]
\label{eq:SEkernel}
\end{equation}
where $\ell_1$, $\ell_2$, and $\ell_3$ are the characteristic length scales associated with $\varepsilon_v$, $\varepsilon_s$, and $p$, respectively, and $\sigma_f^2$ is the signal variance.

\paragraph{Hyperparameter estimation.}
The model hyperparameters
$\boldsymbol{\theta}=\{\ell_1,\ell_2,\ell_3,\sigma_f,\sigma_n\}$
are optimized by minimizing the negative log marginal likelihood, analogous to Eq. \eqref{eq:GPR_marglik}:
\begin{equation}
\begin{aligned}
\mathcal{L}(\boldsymbol{\theta})
= & \frac{1}{2}
\big[\boldsymbol{\Gamma}-m(\mathbf{X})\big]^\top
\big[K_{\boldsymbol{\theta}}+\sigma_n^2 I\big]^{-1}
\big[\boldsymbol{\Gamma}-m(\mathbf{X})\big]
+ \frac{1}{2}\log\big[K_{\boldsymbol{\theta}}+\sigma_n^2 I\big]
+ \frac{N}{2}\log(2\pi),\\ & \quad\text{s.t.}\quad
\mathcal{C}_1(\boldsymbol{\theta})\ge0,\ 
\mathcal{C}_2(\boldsymbol{\theta})\ge0.
\end{aligned}
\label{eq:negloglik}
\end{equation}
where $K_{\boldsymbol{\theta}}=K(\mathbf{X},\mathbf{X};\boldsymbol{\theta})$.  
The key difference is that this optimization is performed subject to the physical constraints, $\mathcal{C}_1\ge0$ and $\mathcal{C}_2\ge0$, which are described below.

\paragraph{Polynomial mean function.}
We first introduce a physics-guided polynomial mean function defined by:
\begin{equation}
\mu(\mathbf{x}) = \mathbf{h}(\mathbf{x})^\top \boldsymbol{\beta},
\label{eq:PI_GPR_mean}
\end{equation}
where $\mathbf{h}(\mathbf{x})$ denotes the vector of polynomial basis functions of total degree $d$, and $\boldsymbol{\beta}$ are the coefficients obtained by solving a constrained ridge least-squares problem:
\begin{equation}
\min_{\boldsymbol{\beta}}
\ \|\mathbf{H}\boldsymbol{\beta}-\boldsymbol{\Gamma}\|_2^2
+ \lambda_{\mathrm{reg}}\|\boldsymbol{\beta}\|_2^2
\quad\text{s.t.}\quad
\mathcal{C}_1(\boldsymbol{\beta})\ge0,\ 
\mathcal{C}_2(\boldsymbol{\beta})\ge0.
\label{eq:PI_GPR_beta}
\end{equation}
Here, $\mathbf{H}$ is the design matrix formed by evaluating $\mathbf{h}(\mathbf{x})$ at the training points and $\lambda_{\mathrm{reg}}$ is a small regularization coefficient.  
Again, the two inequality constraints $\mathcal{C}_1$ and $\mathcal{C}_2$ represent the physical relations introduced below. A more general form of this mean function may follow the constrained polynomial chaos expansion (PC$^2$)~\cite{novak2024physics,sharma2024physics}. 

\paragraph{Constraint $\mathcal{C}_1$: $\Gamma$-hardening with pressure.}
The first constraint addresses the evolution of $\Gamma$ with confining pressure. 
In the KCC model, $\Gamma$ defines the failure surface amplitude, which governs the deviatoric stress level required to reach yielding under a given pressure state. 
Physically, for pressure-sensitive and frictional materials such as concrete, increasing confinement suppresses crack propagation and microcrack opening, resulting in higher overall strength. 
This behavior has been consistently observed in triaxial compression experiments, where both the peak and residual strengths increase with confinement pressure \cite{sfer2002study, imran2001plasticity,palaniswamy1974fracture}. 
Accordingly, the yield surface should expand monotonically (i.e., the material hardens) as pressure increases.
Therefore, to ensure that the failure surface expands monotonically with increasing confinement pressure, the model must satisfy
\begin{equation}
\mathcal{C}_1\coloneq \frac{\partial \Gamma}{\partial p} \ge 0.
\label{eq:constraint_pressure}
\end{equation}
Failure to satisfy this condition would imply that concrete weakens under confinement, leading to nonphysical predictions such as negative dissipation. 
Thus, the $\Gamma$-hardening constraint directly encodes a fundamental physical mechanism, pressure hardening of concrete, into the GPR model, ensuring that the learned surrogate remains consistent with the known effect of confinement on concrete.

This condition is enforced both in the polynomial mean (Eq.~\eqref{eq:PI_GPR_mean}) and in the GP posterior (via Eq.~\eqref{eq:negloglik}) through a probabilistic constraint. 
Following the constrained GPR formulation proposed by Pensoneault \cite{pensoneault2020nonnegativity}, the constraint is implemented by restricting the feasible functional space of the GP through hyperparameter optimization. Instead of a continuous constraint, enforcement is applied at a finite set of virtual derivative points $\{\mathbf{x}_v\}$ sampled from the input domain $\mathcal{X}$. 
At these points, the probability of violating the $\Gamma$-hardening constraint is required to remain small:
\begin{equation}
\mathbb{P}\!\left[
\frac{\partial \Gamma(\mathbf{x}_v)}{\partial p} < 0
\;\middle|\;
\mathbf{x}_v,\boldsymbol{\Gamma},\mathbf{X}
\right]
\le \eta,
\quad \forall\,\mathbf{x}_v\in\mathcal{X},
\label{eq:prob_constraint}
\end{equation}
where $0<\eta\ll1$ denotes the allowable violation probability.  
Adopting the method of constrained likelihood with derivative information in \cite{swiler2020survey}, $\partial \Gamma / \partial p$ is a Gaussian process, and Eq.~\eqref{eq:prob_constraint} can be equivalently expressed in deterministic form as
\begin{equation}
\mu_\frac{\partial \Gamma}{\partial p}
- \Phi^{-1}(\eta)\,
\sigma_\frac{\partial \Gamma}{\partial p}
\ge 0,
\label{eq:det_constraint}
\end{equation}
where $\mu_\frac{\partial \Gamma}{\partial p}$ and $\sigma_\frac{\partial \Gamma}{\partial p}$ are the predicted mean and standard deviation of the derivative $\partial \Gamma/\partial p$, and $\Phi^{-1}$ denotes the inverse standard normal cumulative distribution function. 
With \( p \equiv x_3 \), and according to the polynomial mean function defined in Eq.~\eqref{eq:PI_GPR_mean},  
the derivative of the mean with respect to pressure \( p \) is obtained as  
\begin{equation}
\mu_\frac{\partial \Gamma}{\partial p} = 
\Big(\frac{\partial \mathbf{h}}{\partial x_3}\Big)^\top \boldsymbol{\beta}.
\end{equation}
Analogous to Eq.~\eqref{eq:post_var}, the variance of the predicted derivative \(\partial \Gamma / \partial p\) at any point \(\mathbf{x}_*\) is expressed as
\begin{equation}
\sigma_{\frac{\partial \Gamma}{\partial p}}^2(\mathbf{x}_*)
=
K'(\mathbf{x}_*,\mathbf{x}_*)
-
K'(\mathbf{x}_*,\mathbf{X})
K'(\mathbf{X},\mathbf{X})^{-1}
K'(\mathbf{X},\mathbf{x}_*),
\label{eq:deriv_var_general_updated}
\end{equation}
where \(K'(\mathbf{x},\mathbf{x}')\) denotes the derivative kernel with respect to \(p\) and \(p'\).  
Based on the squared-exponential kernel \(K(\mathbf{x},\mathbf{x}')\) defined in Eq.~\eqref{eq:SEkernel},  
the derivative kernel \(K'(\mathbf{x},\mathbf{x}')\) takes the explicit form
\begin{equation}
K'(\mathbf{x},\mathbf{x}')
\;=\;
\frac{\partial^2 K(\mathbf{x},\mathbf{x}')}{\partial p\,\partial p'}
=
\sigma_f^2
\exp\!\left[-\sum_{j=1}^{3}\frac{(x_j - x'_j)^2}{2\ell_j^2}\right]
\left(
\frac{1}{\ell_3^2}
- \frac{(x_3 - x_3')^2}{\ell_3^4}
\right).
\label{eq:d2K_dpdpprime_full}
\end{equation}
Minimizing Eq.~\eqref{eq:negloglik} subject to Eq.~\eqref{eq:det_constraint} yields hyperparameters that ensure $\partial \Gamma/\partial p \ge 0$ with high probability, thus enforcing physically consistent strength expansion with increased confinement.

\paragraph{Constraint $\mathcal{C}_2$: Regions of $\Gamma$ monotonicity with deviatoric strain.}
The second constraint addresses the evolution of $\Gamma$ with respect to the deviatoric strain $\varepsilon_s$, which reflects the material response under increasing shear deformation. Prior to the failure surface $\Gamma$ reaching its peak value at a critical deviatoric strain $\varepsilon_s^{\max}$, the failure surface will monotonically expand. 
After $\Gamma$ reaches its peak at $\varepsilon_s^{\max}$, the progressive coalescence and localization of microcracks leads to stiffness degradation and softening \cite{markeset1995softening}, during which $\Gamma$ should decrease with further shear deformation. 
At very high confining pressures, or under extreme impact loading where both strain rates and pressures become very large, concrete may exhibit rehardening behavior due to material densification and pore collapse \cite{ren2008behavior,cusatis2011lattice,cusatis2011strain}. 
However, we restrict our attention in this study to conditions where re-hardening can be neglected.
Within the moderate pressure regime relevant to structural applications, 
a deviatoric monotonicity constraint is introduced to suppress nonphysical rehardening in $\Gamma$, while maintaining the physically meaningful post-peak softening behavior characteristic of quasi-brittle materials.

The second constraint is therefore expressed as
\begin{equation}
\mathcal{C}_2 = 
\begin{cases}
\frac{\partial \Gamma}{\partial \varepsilon_s} \ge 0, & \varepsilon_s \le \varepsilon_s^{\max}, \\[4pt]
\frac{\partial \Gamma}{\partial \varepsilon_s} \le 0, & \varepsilon_s > \varepsilon_s^{\max},
\end{cases}
\label{eq:constraint_strain}
\end{equation}
where $\varepsilon_s^{\max}$ again denotes the deviatoric strain at which $\Gamma$ attains its peak.  
Unlike the $\Gamma$-hardening constraint, the deviatoric monotonicity constraint is applied only to the mean function $\mu(\mathbf{x})$ during the constrained optimization in Eq.~\eqref{eq:PI_GPR_beta}. To illustrate its necessity,~\ref{appendix:example_plot} presents an example GPR prediction demonstrating that enforcing only the $\Gamma$-hardening constraint (without the deviatoric monotonicity constraint) can lead to nonphysical post-peak rehardening in $\Gamma$.

Together, these two constraints embed essential physical requirements into the GP framework, ensuring that the learned surrogate $\Gamma(\varepsilon_v,\varepsilon_s,p)$ remains mechanically admissible. It is worth noting that the present framework is not limited to these two forms of physical regularization. In particular, additional constraints may be derived from thermodynamic consistency. For example, in~\ref{appendix:thermo}, we outline how the second law of thermodynamics, expressed through the Clausius--Duhem inequality, leads to conditions relating $\Gamma$ and its pressure sensitivity. Although such thermodynamic constraints are not enforced in the current implementation, the framework is readily extendable to incorporate them in future developments.

\section{Proposed Constitutive Model Training and Performance}
\label{sec:demonstration}

\subsection{Training data preparation}

To demonstrate the proposed constrained GPR constitutive model, we employ experimental data from triaxial compression tests on WSMR\textendash5 concrete \cite{joy1993material}. 
The experimental dataset consists of axisymmetric triaxial tests performed under four different confinement levels of $P_c=7, 14, 20,$ and \SI{34}{MPa}. 
In this study, the deviatoric stress is computed directly from the measured axial and radial stresses, while the input features, volumetric strain, deviatoric strain, and pressure are derived from the recorded strain and stress components. 
These quantities constitute the input–output pairs used for training the GPR surrogate.

The WSMR\textendash5 concrete dataset has been widely used as a benchmark for calibrating the default failure surface parameters of the KCC model \cite{wu2012performance,crawford2012use}. 
To further investigate the influence of data availability on model performance, we also generate synthetic data using the calibrated KCC parameters. 
A series of single-element triaxial compression simulations are conducted over a range of confinement pressures, from which the same input variables (\(\varepsilon_v,\varepsilon_s,p\)) and the output variable ($\Gamma$) are extracted. 
These simulation results serve two purposes. First, since the model has already been well calibrated for the WSMR\textendash5 data set, they provide reference strain-stress response to quantitatively evaluate the accuracy of the proposed GPR model in conditions where experimental data are unavailable. Second, some of the simulation data are incorporated into the training set to expand coverage across the pressure domain, enabling us to examine how model performance improves as more training data become available.

In the subsequent sections, we systematically examine the performance of the proposed material model under different training conditions. We begin by showing how the model performs if the physical constraints are not included, considering the case where only the experimental data are available for training as well as the case where both experimental and simulation data are used for training. We then compare this to the performance of the model with the physical constraints included to demonstrate its enhanced performance. 

\subsection{Unconstrained model performance}

\subsubsection{Experimental Training Data only}

The failure surface function $\Gamma$ is first modeled using the squared-exponential covariance kernel defined in Eq.~\eqref{eq:SEkernel}. 
The posterior mean and variance are computed from Eqs.~\eqref{eq:post_mean}–\eqref{eq:post_var}, and the hyperparameters are optimized using the COBYLA algorithm implemented in the \texttt{SciPy} package \cite{virtanen2020scipy} without including constraints. 
Figure~\ref{Train4_Reg} presents representative deviatoric stress vs. deviatoric strain ($\Gamma$–$\varepsilon_s$) curves obtained when the model is trained solely on the four experimental datasets. 
For clarity, only selected confinement levels are shown in the figure, spanning cases that include both interpolation and extrapolation behavior. A complete set of $\Gamma$–$\varepsilon_s$ curves covering all confinement pressures from $P_c = 5$ to \SI{39}{MPa} is provided in~\ref{appendix:complete set}.
The plots highlighted in orange correspond to the training data and all others show comparison with the calibrated KCC model as a test set. 
In each plot, the red curves represent the GPR posterior mean predictions, and the shaded blue regions indicate the $95\%$ confidence intervals ($\mu_\Gamma \pm 1.96\sigma_\Gamma$) from the unconstrained GPR. 
The blue dashed lines correspond to the reference (“ground-truth”) values, which are either obtained from experimental measurements (for training pressures) or the calibrated KCC model simulations (for testing pressures). 
\begin{figure}[!ht]
\centering
\includegraphics[width=1\linewidth]{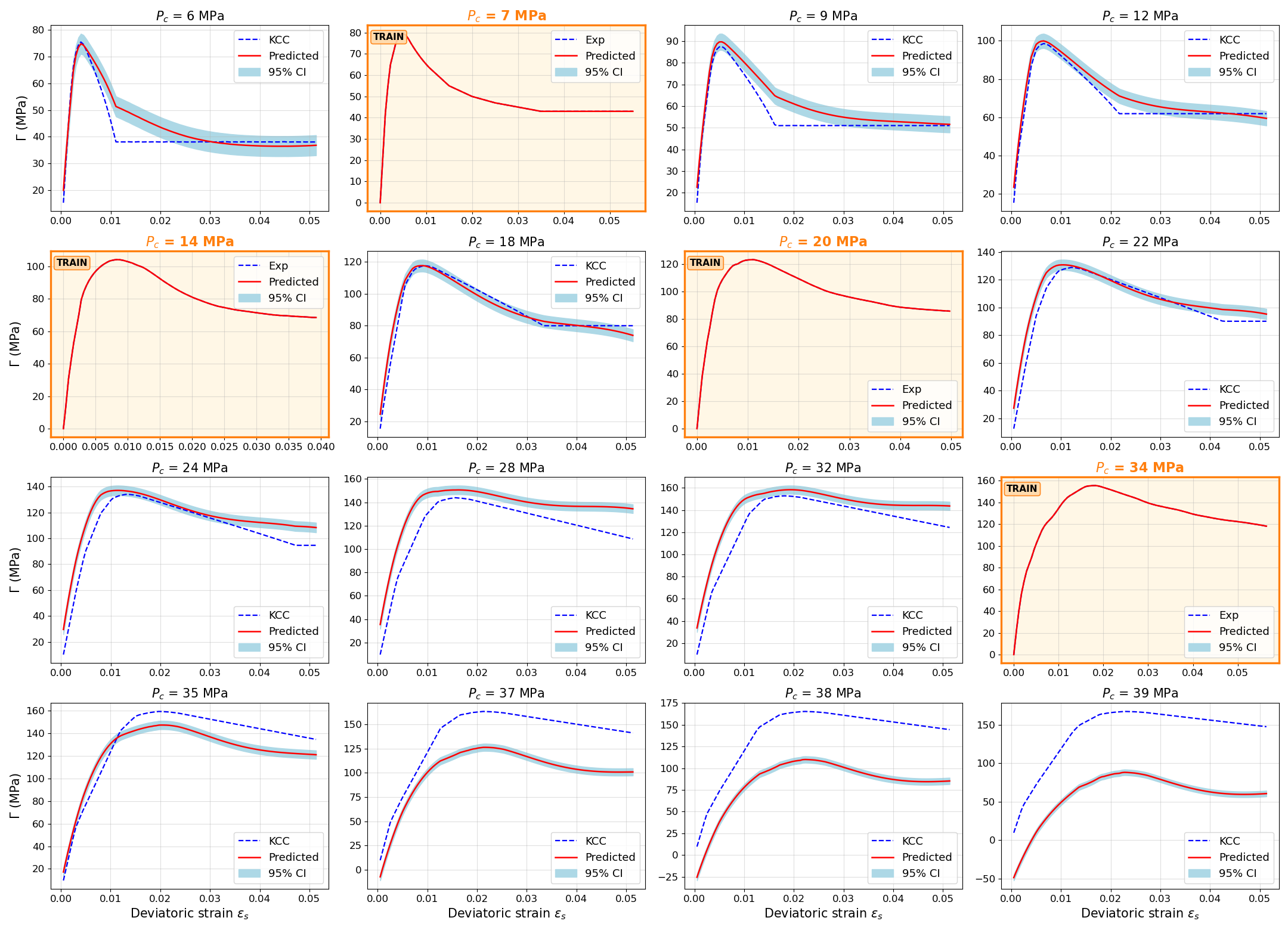}
\caption{Unconstrained GPR constitutive model trained using four experimental datasets showing $\Gamma$–$\varepsilon_s$ relations at representative confinement levels in between 5 to \SI{39}{MPa}. Experimental training data are highlighted in orange.}
\label{Train4_Reg_selected}
\end{figure}

Overall, the unconstrained model reproduces the general trend of $\Gamma-\varepsilon_s$ reasonably well near the training pressures. 
However, the accuracy decreases rapidly as the confinement level deviates from the training data. 
Quantitative comparisons are summarized in Table~\ref{Table} using the normalized root-mean-square error (NRMSE) and the coefficient of determination ($R^2$) for the GPR prediction compared with the KCC model where the second column shows data using 4 experimental training data and the unconstrained GPR.

To facilitate interpretation, we define four accuracy tiers and color code the results in Table~\ref{Table} accordingly: 
(1) \emph{Excellent:} [Green] $\mathrm{NRMSE}<2\%$ and $R^2>0.98$.  
(2) \emph{Good:} [Yellow] $\mathrm{NRMSE}=2\text{–}5\%$ and $R^2=0.85\text{–}0.98$;  
(3) \emph{Acceptable:} [Orange] $\mathrm{NRMSE}=5\text{–}12\%$ and $R^2=0.7\text{–}0.85$;  
(4) \emph{Poor:} [Red] $\mathrm{NRMSE}>12\%$ and $R^2<0.7$. While predictions near the experimentally observed confinement levels (7–20 MPa) fall within the acceptable or good range, substantial deterioration occurs at higher pressures (e.g., around 28 MPa and above), where training data are absent.
This limitation becomes most evident beyond 34 MPa, where the predicted $\Gamma$ erroneously decreases with increasing pressure (see Figure~\ref{Train4_Reg_selected}), contradicting the expected confinement-hardening behavior of concrete.  
The last two rows in Table~\ref{Table} highlight this pronounced degradation, underscoring the restricted generalizability of the unconstrained GPR. Because the unconstrained model lacks any mechanism to enforce the $\Gamma$-hardening constraint $\mathcal{C}_1$, its extrapolation outside the training regime becomes non-physical.

\begin{figure}[!ht]
\centering
\includegraphics[width=1\linewidth]{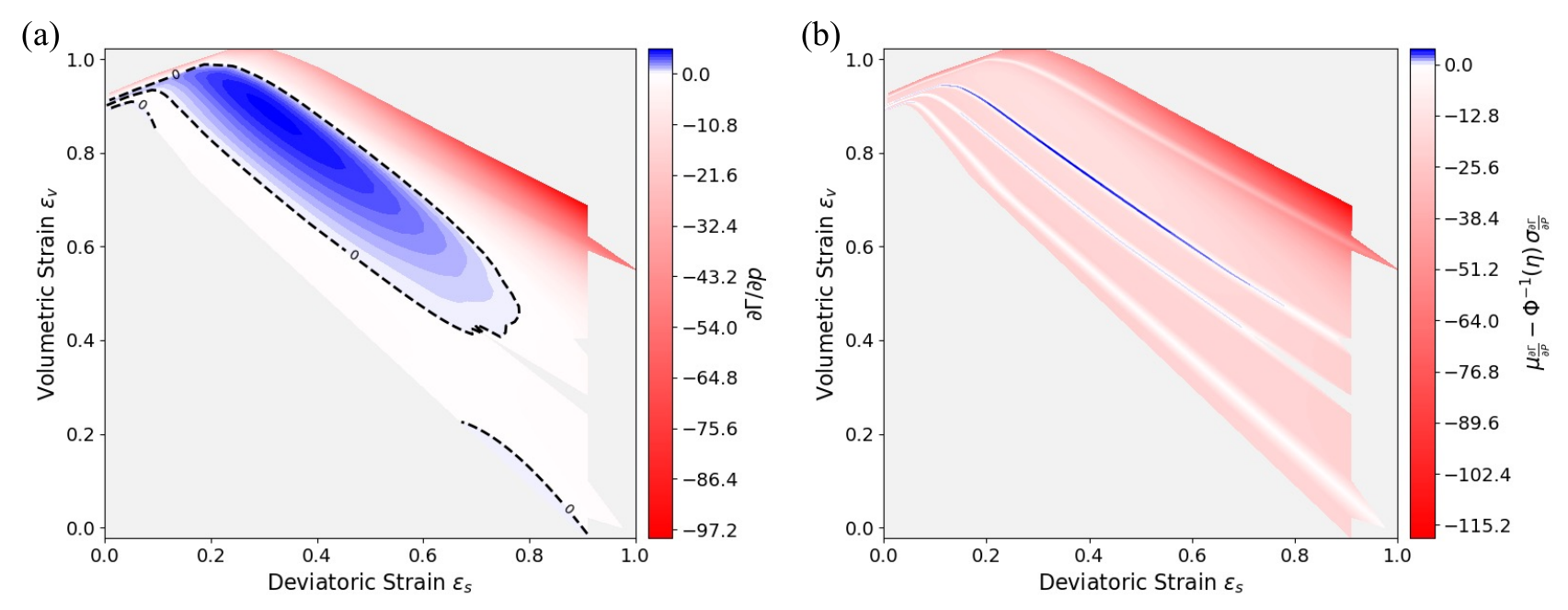}
\caption{Illustration of when the $\Gamma$-hardening constraint $\mathcal{C}_1$ is violated in the unconstrained GPR constitutive model across the $\varepsilon_s$–$\varepsilon_v$ space: (a) mean derivative prediction; (b) 95\% probability violation map considering predictive variance. Blue color shows regions where the constraint is satisfied. Red regions show where the constraint is not satisfied by (a) the mean function and (b) with 95\% confidence. }
\label{Train4_Vio}
\end{figure}

Figure~\ref{Train4_Vio} shows how the unconstrained GP violates the $\Gamma$-hardening constraint $\mathcal{C}_1$ across the $\varepsilon_s$–$\varepsilon_v$ domain. 
In these plots, blue indicates regions where the constraints are satisfied ($\partial\Gamma/\partial p > 0$), red indicates that the constraint is violated, and white corresponds to values of $\partial\Gamma/\partial p$ near zero. 
Because both $\Gamma$ and its derivatives follow Gaussian distributions, the figures distinguish between the mean prediction of $\partial\Gamma/\partial p$ [Fig.~\ref{Train4_Vio}(a)] and the 95\% violation probability that accounts for the predictive variance [Fig.~\ref{Train4_Vio}(b)]. 
It is observed that the constraint is satisfied only in limited regions for the mean prediction. 
However, when the predictive variance is considered, violations appear across nearly the entire input domain. 
This highlights the importance of evaluating both the mean and uncertainty when assessing constraint satisfaction: even if the mean prediction locally satisfies the constraint, a large associated variance implies a high probability of constraint violation. 
Therefore, Figure~\ref{Train4_Vio} demonstrates that without enforcing the $\Gamma$-hardening constraint $\mathcal{C}_1$, the unconstrained GP exhibits a high probablity of widespread violation across the $\varepsilon_s$–$\varepsilon_v$ space.

\begin{table}[!ht]
\centering
\caption{Regression accuracy of the proposed GPR constitutive models under different training data and constraint settings. NRMSE and $R^2$ are reported at various confinement pressures $P_c$. Cells are color coded according to the following measure of fit: (1) \emph{Excellent:} [Green] $\mathrm{NRMSE}<2\%$ and $R^2>0.98$.  
(2) \emph{Good:} [Yellow] $\mathrm{NRMSE}=2\text{–}5\%$ and $R^2=0.85\text{–}0.98$;  
(3) \emph{Acceptable:} [Orange] $\mathrm{NRMSE}=5\text{–}12\%$ and $R^2=0.7\text{–}0.85$;  
(4) \emph{Poor:} [Red] $\mathrm{NRMSE}>12\%$ and $R^2<0.7$. }
\begin{tabular}{c|cc|cc|cc|cc}
\hline
& \multicolumn{6}{c|}{Unconstrained} &
\multicolumn{2}{c}{Constrained} \\
\hline
\multirow{2}{*}{$P_c$ (MPa)} &
\multicolumn{2}{c|}{4 Data} &
\multicolumn{2}{c|}{8 Data} &
\multicolumn{2}{c|}{12 Data} &
\multicolumn{2}{c}{4 Data} \\
 & NRMSE & $R^2$ & NRMSE & $R^2$ & NRMSE & $R^2$ & NRMSE & $R^2$ \\ \hline
6  & \cellcolor{orange!50!white} 8.32\%  & \cellcolor{orange!50!white} 0.7802 & \cellcolor{yellow!50} 3.42\%  & \cellcolor{yellow!50} 0.9629 & \cellcolor{yellow!50} 3.44\%  & \cellcolor{yellow!50} 0.9590 & \cellcolor{orange!50} 11.16\% & \cellcolor{red!50} 0.6047 \\ \hline
8  & \cellcolor{orange!50} 8.74\%  & \cellcolor{orange!50} 0.7450 & \cellcolor{yellow!50} 3.60\%  & \cellcolor{yellow!50} 0.9568 & \cellcolor{yellow!50} 3.59\%  & \cellcolor{yellow!50} 0.9548 & \cellcolor{orange!50} 7.74\%  & \cellcolor{orange!50} 0.8001 \\ \hline
9  & \cellcolor{orange!50} 8.13\%  & \cellcolor{orange!50} 0.7764 & \cellcolor{yellow!50} 3.55\%  & \cellcolor{yellow!50} 0.9573 & \cellcolor{yellow!50} 3.55\%  & \cellcolor{yellow!50} 0.9563 & \cellcolor{orange!50} 6.64\%  & \cellcolor{yellow!50} 0.8505 \\ \hline
11 & \cellcolor{orange!50} 6.02\%  & \cellcolor{yellow!50} 0.8765 & \cellcolor{yellow!50} 3.49\%  & \cellcolor{yellow!50} 0.9584 & \cellcolor{yellow!50} 3.44\%  & \cellcolor{yellow!50} 0.9592 & \cellcolor{yellow!50} 4.77\%  & \cellcolor{yellow!50} 0.9223 \\ \hline
12 & \cellcolor{yellow!50} 4.93\%  & \cellcolor{yellow!50} 0.9178 & \cellcolor{yellow!50} 3.45\%  & \cellcolor{yellow!50} 0.9599 & \cellcolor{yellow!50} 3.35\%  & \cellcolor{yellow!50} 0.9613 & \cellcolor{yellow!50} 4.01\%  & \cellcolor{yellow!50} 0.9458 \\ \hline
13 & \cellcolor{yellow!50} 4.05\%  & \cellcolor{yellow!50} 0.9451 & \cellcolor{yellow!50} 3.34\%  & \cellcolor{yellow!50} 0.9628 & \cellcolor{yellow!50} 3.15\%  & \cellcolor{yellow!50} 0.9651 & \cellcolor{yellow!50} 3.53\%  & \cellcolor{yellow!50} 0.9585 \\ \hline
15 & \cellcolor{yellow!50} 3.56\%  & \cellcolor{yellow!50} 0.9595 & \cellcolor{yellow!50} 3.30\%  & \cellcolor{yellow!50} 0.9651 & \cellcolor{yellow!50} 2.65\%  & \cellcolor{yellow!50} 0.9744 & \cellcolor{yellow!50} 3.42\%  & \cellcolor{yellow!50} 0.9626 \\ \hline
16 & \cellcolor{yellow!50} 3.84\%  & \cellcolor{yellow!50} 0.9540 & \cellcolor{yellow!50} 3.65\%  & \cellcolor{yellow!50} 0.9584 & \cellcolor{yellow!50} 2.57\%  & \cellcolor{yellow!50} 0.9753 & \cellcolor{yellow!50} 3.39\%  & \cellcolor{yellow!50} 0.9641 \\ \hline
18 & \cellcolor{yellow!50} 4.13\%  & \cellcolor{yellow!50} 0.9498 & \cellcolor{yellow!50} 3.37\%  & \cellcolor{yellow!50} 0.9667 & \cellcolor{yellow!50} 2.22\%  & \cellcolor{green!50} 0.9838 & \cellcolor{yellow!50} 3.89\%  & \cellcolor{yellow!50} 0.9556 \\ \hline
19 & \cellcolor{yellow!50} 4.36\%  & \cellcolor{yellow!50} 0.9454 & \cellcolor{yellow!50} 3.05\%  & \cellcolor{yellow!50} 0.9733 & \cellcolor{yellow!50} 2.02\%  & \cellcolor{green!50} 0.9872 & \cellcolor{yellow!50} 4.21\%  & \cellcolor{yellow!50} 0.9490 \\ \hline
21 & \cellcolor{orange!50} 5.51\%  & \cellcolor{yellow!50} 0.9157 & \cellcolor{yellow!50} 2.55\%  & \cellcolor{green!50} 0.9820 & \cellcolor{green!50} 1.54\%  & \cellcolor{green!50} 0.9931 & \cellcolor{yellow!50} 4.74\%  & \cellcolor{yellow!50} 0.9375 \\ \hline
22 & \cellcolor{orange!50} 6.46\%  & \cellcolor{yellow!50} 0.8854 & \cellcolor{yellow!50} 2.26\%  & \cellcolor{green!50} 0.9860 & \cellcolor{green!50} 1.25\%  & \cellcolor{green!50} 0.9955 & \cellcolor{orange!50} 5.08\%  & \cellcolor{yellow!50} 0.9290 \\ \hline
24 & \cellcolor{orange!50} 8.87\%  & \cellcolor{orange!50} 0.7899 & \cellcolor{green!50} 1.95\%  & \cellcolor{green!50} 0.9899 & \cellcolor{green!50} 0.81\%  & \cellcolor{green!50} 0.9982 & \cellcolor{orange!50} 5.77\%  & \cellcolor{yellow!50} 0.9110 \\ \hline
25 & \cellcolor{orange!50} 10.08\% & \cellcolor{orange!50} 0.7324 & \cellcolor{green!50} 1.93\%  & \cellcolor{green!50} 0.9901 & \cellcolor{green!50} 0.73\%  & \cellcolor{green!50} 0.9986 & \cellcolor{orange!50} 6.03\%  & \cellcolor{yellow!50} 0.9043 \\ \hline
26 & \cellcolor{orange!50} 11.35\% & \cellcolor{red!50} 0.6653 & \cellcolor{green!50} 1.98\%  & \cellcolor{green!50} 0.9898 & \cellcolor{green!50} 0.71\%  & \cellcolor{green!50} 0.9987 & \cellcolor{orange!50} 6.21\%  & \cellcolor{yellow!50} 0.8999 \\ \hline
28 & \cellcolor{red!50} 13.39\% & \cellcolor{red!50} 0.5511 & \cellcolor{yellow!50} 2.51\%  & \cellcolor{green!50} 0.9842 & \cellcolor{green!50} 0.93\%  & \cellcolor{green!50} 0.9978 & \cellcolor{orange!50} 6.66\%  & \cellcolor{yellow!50} 0.8891 \\ \hline
30 & \cellcolor{red!50} 13.78\% & \cellcolor{red!50} 0.5503 & \cellcolor{yellow!50} 3.40\%  & \cellcolor{yellow!50} 0.9727 & \cellcolor{green!50} 1.43\%  & \cellcolor{green!50} 0.9947 & \cellcolor{orange!50} 7.21\%  & \cellcolor{yellow!50} 0.8768 \\ \hline
32 & \cellcolor{orange!50} 10.96\% & \cellcolor{orange!50} 0.7334 & \cellcolor{yellow!50} 4.23\%  & \cellcolor{yellow!50} 0.9602 & \cellcolor{yellow!50} 2.31\%  & \cellcolor{green!50} 0.9861 & \cellcolor{orange!50} 7.90\%  & \cellcolor{yellow!50} 0.8616 \\ \hline
33 & \cellcolor{orange!50} 8.29\%  & \cellcolor{yellow!50} 0.8526 & \cellcolor{yellow!50} 4.52\%  & \cellcolor{yellow!50} 0.9561 & \cellcolor{yellow!50} 2.94\%  & \cellcolor{yellow!50} 0.9773 & \cellcolor{orange!50} 8.35\%  & \cellcolor{yellow!50} 0.8503 \\ \hline
35 & \cellcolor{orange!50} 9.47\%  & \cellcolor{orange!50} 0.8205 & \cellcolor{yellow!50} 3.71\%  & \cellcolor{yellow!50} 0.9725 & \cellcolor{yellow!50} 3.19\%  & \cellcolor{yellow!50} 0.9752 & \cellcolor{orange!50} 9.19\%  & \cellcolor{orange!50} 0.8307 \\ \hline
37 & \cellcolor{red!50} 24.35\% & \cellcolor{red!50} -0.1057 & \cellcolor{yellow!50} 2.86\%  & \cellcolor{green!50} 0.9848 & \cellcolor{green!50} 1.60\%  & \cellcolor{green!50} 0.9953 & \cellcolor{orange!50} 10.39\% & \cellcolor{orange!50} 0.7988 \\ \hline
38 & \cellcolor{red!50} 36.14\% & \cellcolor{red!50} -1.3518 & \cellcolor{orange!50} 7.84\% & \cellcolor{yellow!50} 0.8892 & \cellcolor{yellow!50} 4.54\%  & \cellcolor{yellow!50} 0.9693 & \cellcolor{orange!50} 11.12\% & \cellcolor{orange!50} 0.7774 \\ \hline
39 & \cellcolor{red!50} 51.96\% & \cellcolor{red!50} -3.6955 & \cellcolor{red!50} 15.25\% & \cellcolor{red!50} 0.5955 & \cellcolor{orange!50} 9.44\%  & \cellcolor{yellow!50} 0.8938 & \cellcolor{orange!50} 11.95\% & \cellcolor{orange!50} 0.7514 \\ \hline
\textbf{Mean} & 11.60\% & 0.49 & 3.88\% & 0.95 & 2.67\% & 0.98 & 6.67\% & 0.88 \\ \hline
\end{tabular}
\label{Table}
\end{table}

\subsubsection{Augmented Experimental and Simulation Training Data}

\begin{figure}[!ht]
\centering
\includegraphics[width=1\linewidth]{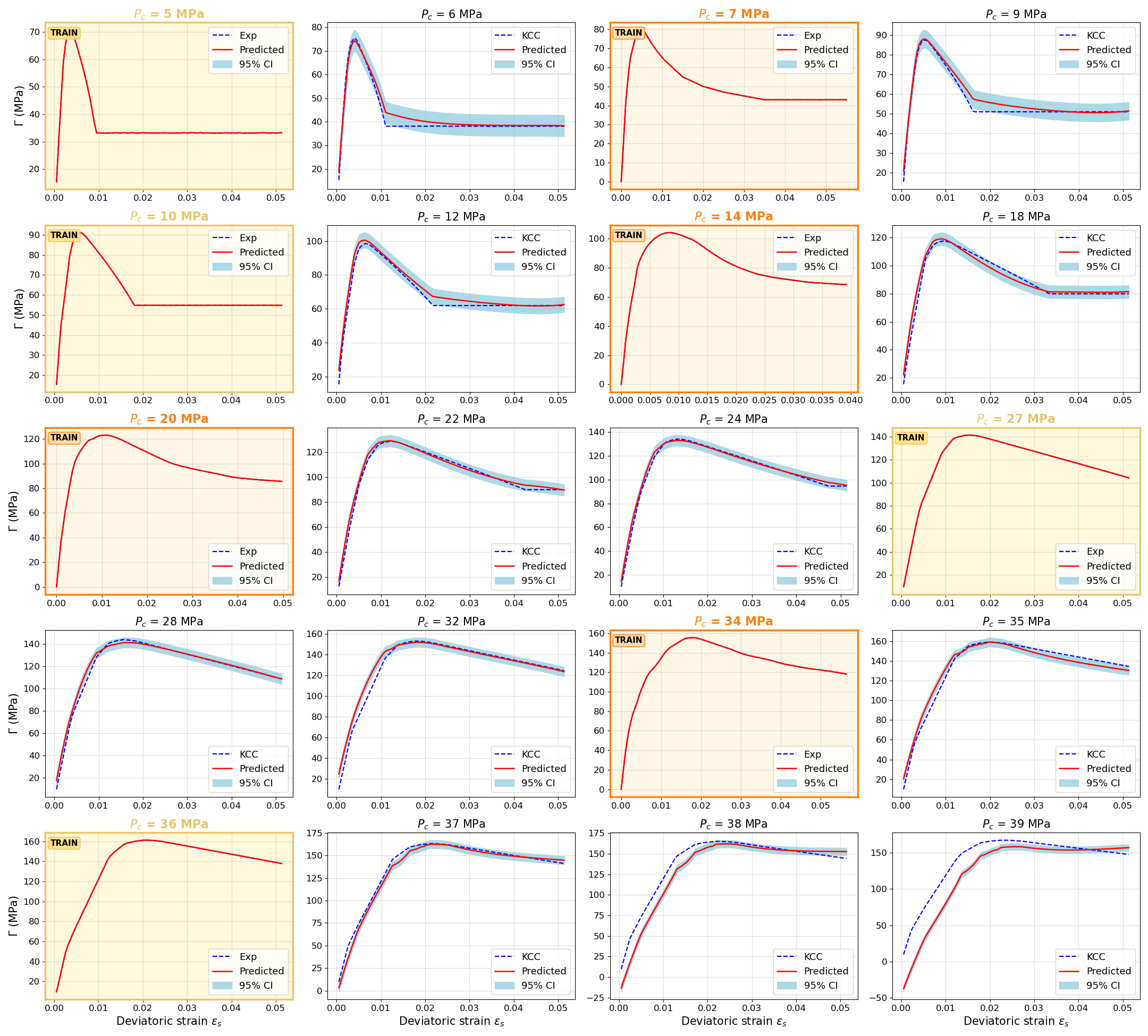}
\caption{Unconstrained GPR constitutive model trained using four experimental and four simulation datasets showing $\Gamma$–$\varepsilon_s$ relations at representative confinement levels in between 5 to \SI{39}{MPa}. Experimental training data are highlighted in orange and simulated training datasets are highlighted in yellow.}
\label{Train8_Reg_selected}
\end{figure}

\begin{figure}[!ht]
\centering
\includegraphics[width=1\linewidth]{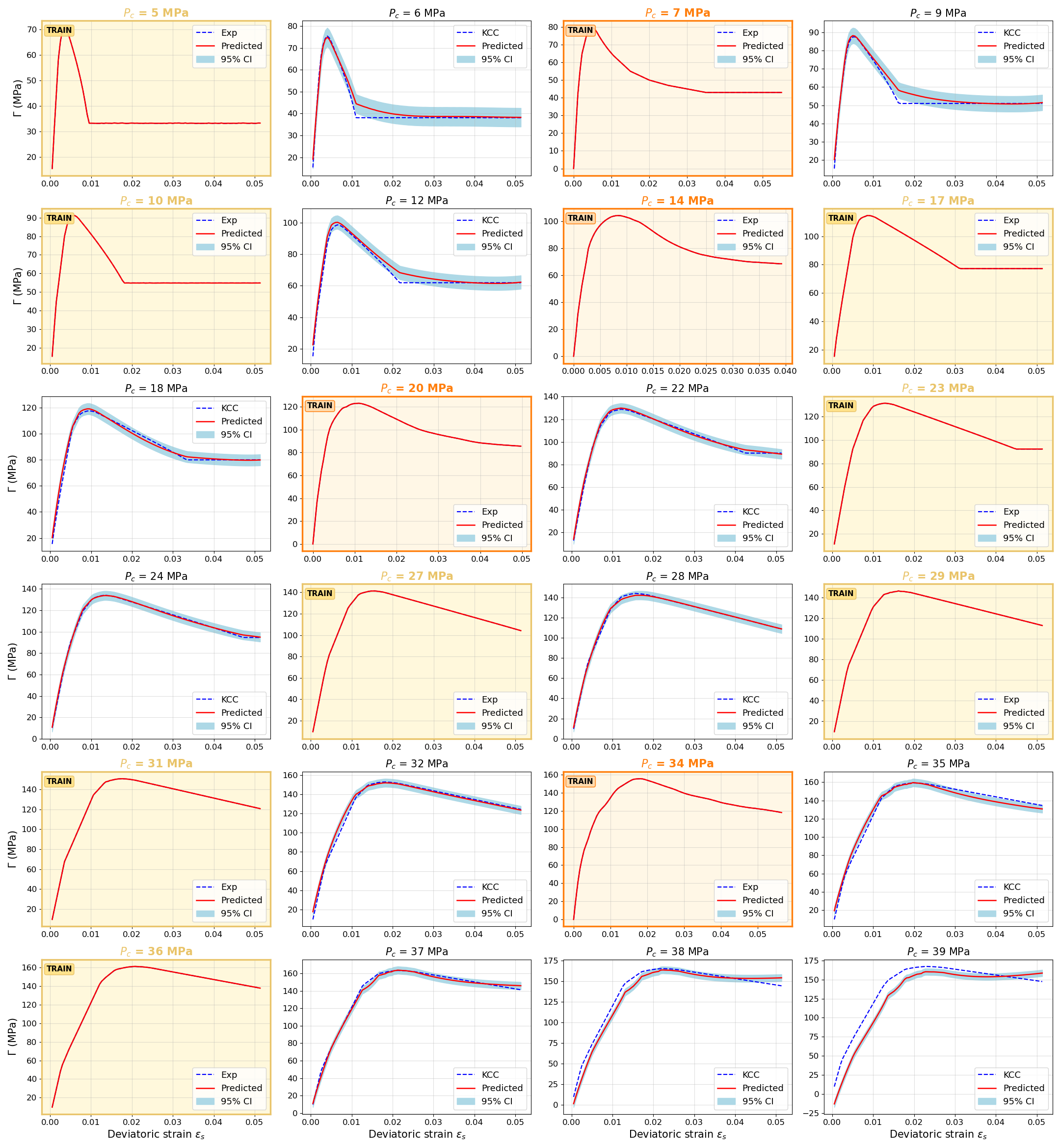}
\caption{Unconstrained GPR constitutive model trained using four experimental and eight simulation datasets showing $\Gamma$–$\varepsilon_s$ relations at representative confinement levels in between 5 to \SI{39}{MPa}. Experimental training data are highlighted in orange and simulated training datasets are highlighted in yellow.}
\label{Train12_Reg_selected}
\end{figure}

In this section, we examine how increasing the density of training data affects the performance of the unconstrained GP model. 
Two cases are considered. One model is trained using four experimental datasets augmented by four simulated datasets. Another model is trained using four experimental plus eight simulated datasets. 
The simulated data are generated from the calibrated KCC model and serve to enrich the coverage of the strain–pressure space. 
Figures~\ref{Train8_Reg_selected} and \ref{Train12_Reg_selected} present the $\Gamma$–$\varepsilon_s$ relations for these two cases, respectively, with the experimental training datasets are highlighted in orange and simulated training datasets are highlighted in yellow. 
Again, Table~\ref{Table} summarizes the corresponding quantitative metrics, including the NRMSE and the $R^2$ values.

Here we see that the inclusion of four additional simulated datasets substantially improves prediction accuracy compared with the model trained on experimental data alone. 
Across most confinement levels, the results fall within the “good” or “excellent” accuracy ranges defined earlier. 
Further expanding the training set to twelve groups (Fig.~\ref{Train12_Reg_selected}) provides an additional improvement, with many pressure levels achieving very accurate agreement with the reference data. 
These results demonstrate the expected trend that a denser training dataset yields better predictive performance within the data-supported range.

\begin{figure}[!ht]
\centering
\includegraphics[width=1\linewidth]{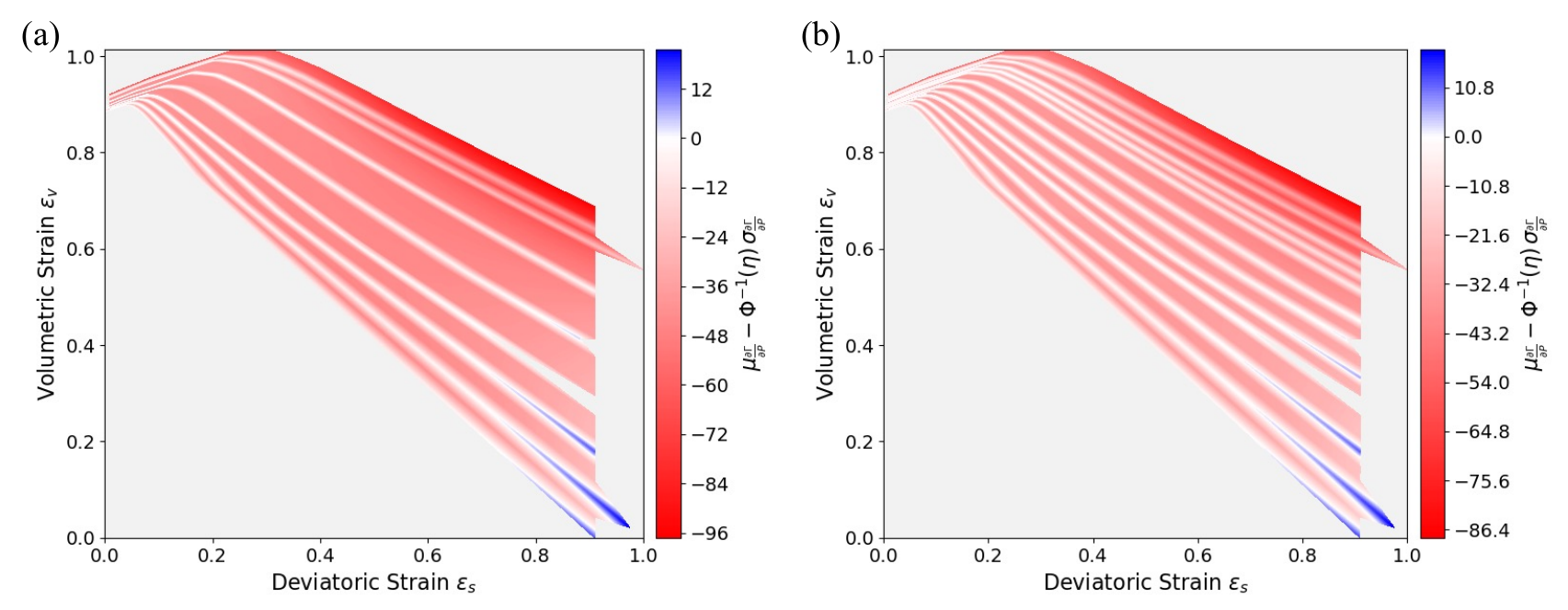}
\caption{95\% confidence $\Gamma$-hardening constraint $\mathcal{C}_1$ for the unconstrained GPR models with increased training data: (a) 4 experimental + 4 simulated data; (b) 4 experimental + 8 simulated data. Blue regions show areas where the model can be guaranteed to satisfy $\mathcal{C}_1$ with 95\% confidence. Red regions show areas where the model cannot be guaranteed to satisfy $\mathcal{C}_1$ with 95\% confidence.}
\label{Train8_12_Vio}
\end{figure}

However, 
the unconstrained GP lacks any mechanism to enforce physically consistent trends — specifically, the monotonic increase of $\Gamma$ with pressure imposed by $\mathcal{C}_1$.
To illustrate this, Figure~\ref{Train8_12_Vio} shows the distribution of the $\Gamma$-hardening constraint $\mathcal{C}_1$ over the $\varepsilon_s$–$\varepsilon_v$ space for the $4+4$-group [Fig.~\ref{Train8_12_Vio}(a)] and $4+8$-group [Fig.~\ref{Train8_12_Vio}(b)] training cases. 
We observed that, while increasing the training data density slightly enlarges the regions where $\partial\Gamma/\partial p > 0$, constraint satisfaction remains confined to neighborhoods immediately surrounding the training data. 
In regions lacking direct data support, the constraint cannot be ensured with high probability. 
This behavior confirms that, without explicit physics-based enforcement, the unconstrained GPR can only interpolate accurately within the training range but fails to generalize beyond it in a physically meaningful way. 
Simply increasing data density improves interpolation but does not guarantee physically meaningful generalization across the full strain–pressure domain.

\subsection{Physics-informed GPR}

Finally, we present the results of the proposed physics-informed constrained GPR constitutive model. 
The two constraints introduced earlier—the $\Gamma$-hardening constraint with respect to pressure and the monotonicity constraints with respect to deviatoric strain—are enforced on the mean function of $\Gamma$. 
The hyperparameters are optimized using the COBYLA algorithm, and the probabilistic constraint in Eq.~\eqref{eq:prob_constraint} is applied with a tolerance level $\eta = 0.025$.

\begin{figure}[!ht]
\centering
\includegraphics[width=1\linewidth]{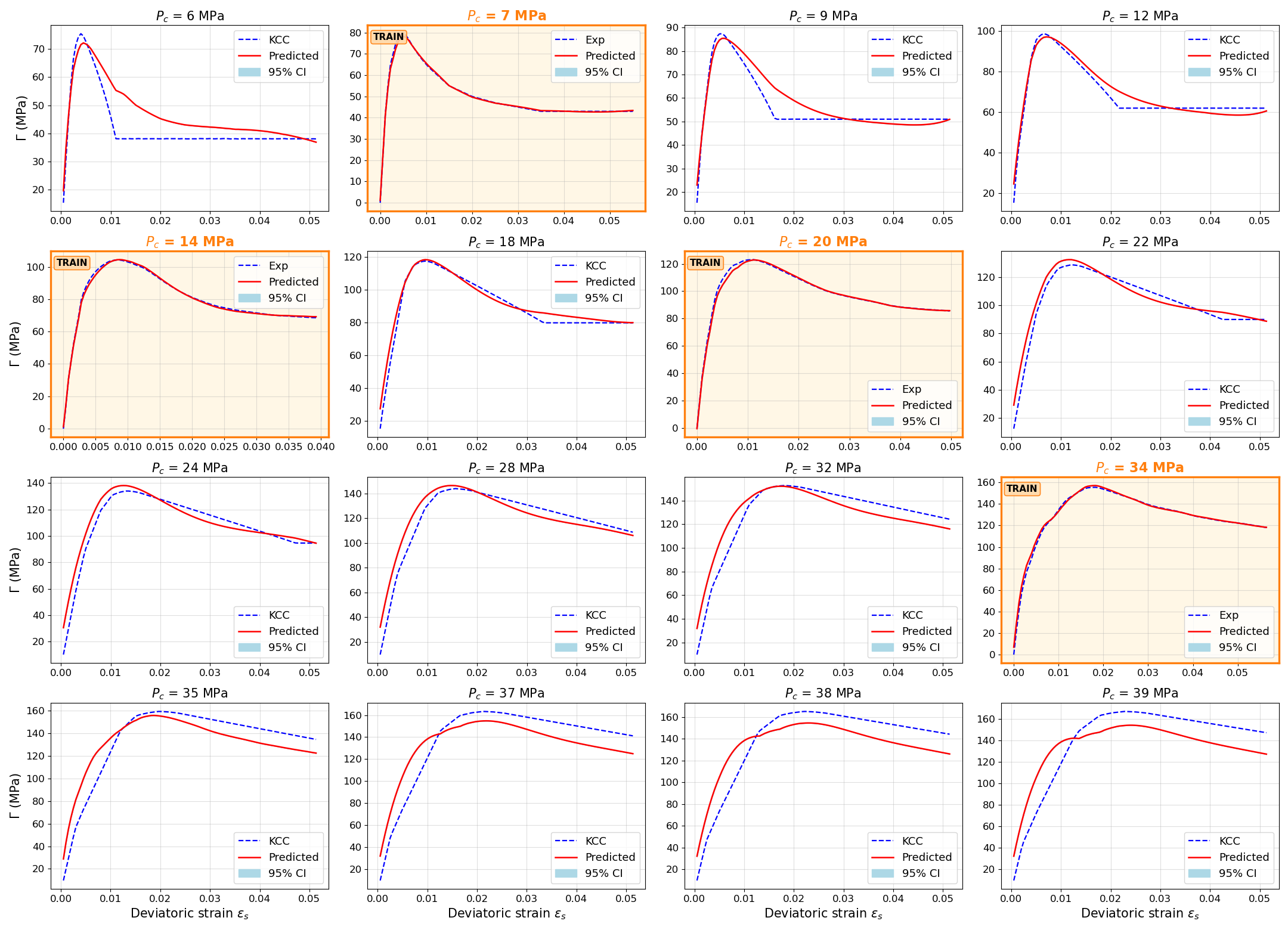}
\caption{Constrained GPR constitutive model trained using four experimental datasets showing $\Gamma$–$\varepsilon_s$ relations at representative confinement levels in between 5 to \SI{39}{MPa}. Experimental training data are highlighted in orange.}
\label{Train4_Con_Reg_selected}
\end{figure}

Only the four experimental datasets are used for training in this case, providing a direct comparison with the unconstrained GPR model trained on the same limited data. 
Figure~\ref{Train4_Con_Reg_selected} shows the predicted $\Gamma$–$\varepsilon_s$ relations at different confinement pressures. 
The corresponding quantitative results of NRMSE and $R^2$ are again summarized in Table~\ref{Table}. 
Compared with the unconstrained GPR, the constrained GPR exhibits a significant improvement in prediction accuracy across nearly all confinement levels that is comparable in accuracy under many conditions to the unconstrained model trained with twice as much data. 
At moderate pressures, the results achieve “good” agreement with the reference data. 
More importantly, even at high pressures ($P_c = 38$–\SI{39}{MPa}), which lie beyond the training data range, the constrained GPR maintains reliable predictions with notably lower error than the unconstrained case. 
This demonstrates the strong benefit of physics-informed regularization, which improves extrapolation quality without the need for additional data.

Another key advantage of the constrained formulation lies in its improved uncertainty behavior. 
By enforcing the probabilistic constraint in Eq.~\eqref{eq:prob_constraint}, the variance of the GP posterior is effectively bounded within the physically admissible region, leading to much tighter confidence intervals and more stable predictions. 
This feature is especially beneficial in data-scarce regimes, where uncontrolled variance can otherwise result in unreliable or nonphysical extrapolations.

\begin{figure}[!ht]
\centering
\includegraphics[width=1\linewidth]{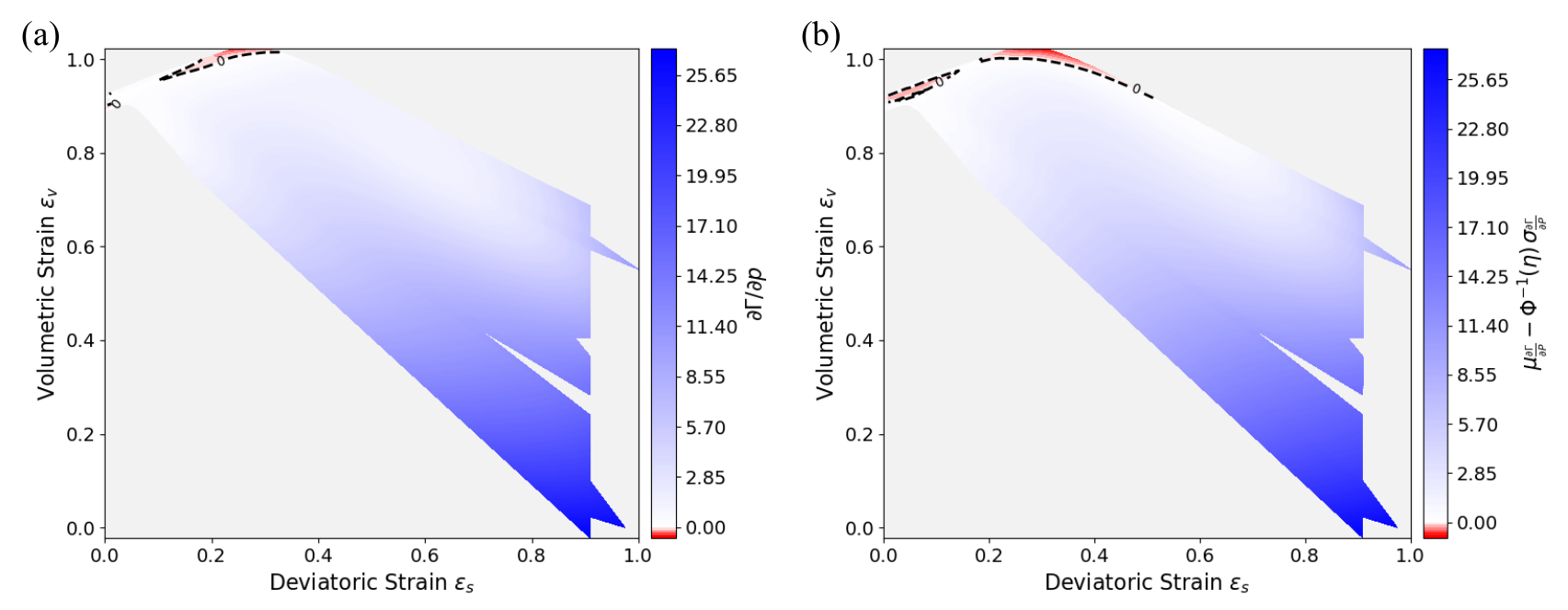}
\caption{Illustration of the $\Gamma$-hardening constraint $\mathcal{C}_1$ in the unconstrained GPR constitituve model across the $\varepsilon_s$–$\varepsilon_v$ space: (a) mean derivative prediction; (b) 95\% probability violation map considering predictive variance. Blue color shows regions where the constraint is satisfied. Red regions show where the constraint is not satisfied by (a) the mean function and (b) with 95\% confidence. }
\label{Train4_Con_Vio}
\end{figure}

Figure~\ref{Train4_Con_Vio} illustrates the values of $\partial\Gamma/\partial p$ for the constrained GPR. 
In Fig.~\ref{Train4_Con_Vio}(a), the mean prediction remains positive throughout nearly the entire $\varepsilon_s$–$\varepsilon_v$ space, confirming successful enforcement of the $\Gamma$-hardening constraint $\mathcal{C}_1$. 
When predictive variance is taken into account [Fig.~\ref{Train4_Con_Vio}(b)], the constraint remains satisfied with 95\% confidence in almost all regions, indicating that the probabilistic enforcement effectively preserves physical consistency even under uncertainty. 
In contrast to the widespread violations observed in the unconstrained case, the constrained GPR establishes a globally consistent and physically meaningful relationship between $\Gamma$ and pressure that generalizes across the range of strains and pressures.

Overall, these results demonstrate that the constrained GPR substantially enhances both accuracy and generalizability compared with the unconstrained formulations, even when trained on very small data sets.
By embedding essential physical constraints into the learning process, the model can achieve comparable or even superior predictive performance with fewer training data, offering a practical and data-efficient path toward robust, physics-informed constitutive modeling.

\subsection{Discussion and Implications}
The results presented above highlight a broader implication of this work: carefully selected physical constraints can substantially enhance the reliability of data driven constitutive models, even when only limited experimental data are available. By constraining the learned failure surface to follow fundamental trends observed in concrete, the proposed framework avoids the need for extensive datasets or heavy regularization to control extrapolative behavior. This suggests a practical pathway for integrating machine learning into established constitutive models used in engineering analysis by targeting only calibration sensitive components with data driven surrogates while retaining the overall computational framework. 

In the broader context of constitutive modeling, uncertainty quantification has traditionally relied on forward propagation of uncertain material parameters through deterministic constitutive laws using Monte Carlo or related sampling based techniques. Parameters are typically assigned probability distributions based on calibration data, and thousands of finite element simulations are performed to propagate these uncertainties to macroscopic responses. Although conceptually straightforward, such approaches are computationally demanding and limited in scope, as they account only for parameter variability while neglecting uncertainty in the constitutive form itself. Moreover, these methods provide no mechanism to update or refine uncertainty estimates as new data become available, requiring complete recalibration each time additional experimental information is obtained. In contrast, the Gaussian Process formulation developed in this work provides an intrinsic and analytical form of uncertainty quantification. The predictive variance of the Gaussian Process directly represents epistemic uncertainty arising from data sparsity, naturally expanding in regions with limited observations and contracting where data are dense. When combined with physics-based constraints, this uncertainty is further regularized within physically admissible bounds, preventing nonphysical predictions in extrapolated regions. 
Furthermore, the nonparametric nature of the Gaussian Process allows seamless incorporation of new experimental or simulated data, as each new observation analytically updates the posterior distribution, refining both the mean prediction and associated uncertainty. Together, these features establish a foundation for adaptive, data efficient, and physically consistent uncertainty quantification in constitutive modeling.

\section{Conclusions}

This study developed a physics-informed Gaussian Process Regression (GPR) framework to replace the phenomenological failure surface $\Gamma$ in the Karagozian \& Case concrete (KCC) model with a data-driven surrogate. By embedding two derivative-based constraints—the $\Gamma$-hardening constraint with respect to pressure and a two-part monotonicity constraint with respect to deviatoric strain—the approach preserves the interpretability of the KCC formulation while leveraging the flexibility and uncertainty quantification capabilities of GPR.

Unconstrained GPR, trained solely on limited experimental data, reproduced trends near the training pressures but degraded under extrapolation, particularly at high confinement, sometimes predicting nonphysical decreases of $\Gamma$ with increasing pressure. Augmenting the training set with simulated data improved interpolation accuracy yet did not remedy violations of the pressure-hardening trend outside the immediate training range. In contrast, the constrained GPR maintained high accuracy across a broad set of pressures, including those beyond the training domain, with very limited data and enforced physically consistent behavior by ensuring $\partial\Gamma/\partial p \ge 0$. Probabilistic enforcement further reduced predictive variance, yielding tighter confidence intervals and more reliable predictions in data-scarce settings.

Overall, the results demonstrate that incorporating targeted, physics-based constraints into the GPR learning process substantially enhances data efficiency, physical fidelity, and generalizability of phenomenolgoical material models. The proposed framework provides a practical and interpretable surrogate for the KCC failure surface, integrating experimental observations with principled uncertainty quantification to produce robust predictions across the strain–pressure domain.

\bibliographystyle{unsrt}
\bibliography{reference}
\appendix
\renewcommand\thesection{Appendix \Alph{section}}
\section{Example Demonstration of the Deviatoric Monotonicity Constraint}
\label{appendix:example_plot}
\begin{figure}[!ht]
\centering
\includegraphics[width=0.8\linewidth]{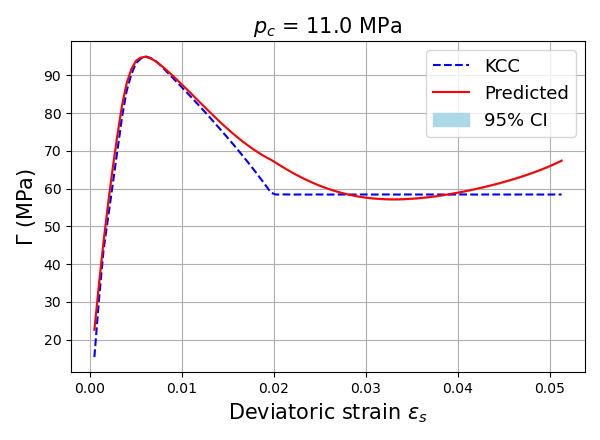}
\caption{Example GPR prediction demonstrating the effect of omitting the deviatoric monotonicity constraint. 
Shown is the evolution of the failure surface $\Gamma$ at a representative confining pressure. 
When only the $\Gamma$-hardening constraint $\mathcal{C}_1$ is enforced, the surrogate exhibits nonphysical post-peak rehardening with respect to $\varepsilon_s$, violating expected quasi-brittle softening behavior.}
\label{fig:rehardening_example}
\end{figure}

Figure~\ref{fig:rehardening_example} shows an example prediction of the trained GPR material model showing that, when only the $\Gamma$-hardening constraint is enforced and the deviatoric monotonicity constraint is omitted, the learned failure surface exhibits nonphysical post-peak rehardening in $\Gamma$.

\section{Thermodynamic Consistency Derivation}
\label{appendix:thermo}

This appendix provides the thermodynamic consistency derivation referenced in Section~2, showing how the Clausius–Duhem inequality leads to a condition relating the failure surface $\Gamma$ and its pressure sensitivity, although this condition is not enforced in the present version of the model.
\subsection*{B.1 From Clausius--Duhem Inequality to Plastic Dissipation}

In an isothermal mechanical process (no temperature gradients and constant temperature), the \textit{Clausius--Duhem inequality} expresses the second law of thermodynamics as a restriction on the rate of mechanical energy dissipation. It states that the rate of internal entropy production must be non-negative:
\begin{equation}
\boldsymbol{\sigma} : \dot{\boldsymbol{\varepsilon}} - \dot{\psi} \ge 0,
\end{equation}
where $\boldsymbol{\sigma}$ is the Cauchy stress tensor, $\dot{\boldsymbol{\varepsilon}}$ is the total strain rate tensor, and $\psi$ is the Helmholtz free energy per unit volume (the stored elastic energy).  
For materials exhibiting elastic--plastic behavior, the total strain rate decomposes as
\begin{equation}
\dot{\boldsymbol{\varepsilon}} = \dot{\boldsymbol{\varepsilon}}^e + \dot{\boldsymbol{\varepsilon}}^p,
\end{equation}
where $\dot{\boldsymbol{\varepsilon}}^e$ and $\dot{\boldsymbol{\varepsilon}}^p$ are the elastic and plastic strain rates, respectively.  
Substituting into the Clausius--Duhem inequality yields
\begin{equation}
\boldsymbol{\sigma} : (\dot{\boldsymbol{\varepsilon}}^e + \dot{\boldsymbol{\varepsilon}}^p) - \dot{\psi} \ge 0.
\end{equation}
Assuming that the free energy $\psi$ depends only on the elastic strain,
\begin{equation}
\psi = \psi(\boldsymbol{\varepsilon}^e),
\end{equation}
the stress is defined as the thermodynamic conjugate variable to the elastic strain:
\begin{equation}
\boldsymbol{\sigma} = \frac{\partial \psi}{\partial \boldsymbol{\varepsilon}^e}.
\end{equation}
Differentiating $\psi$ gives
\begin{equation}
\dot{\psi} = \frac{\partial \psi}{\partial \boldsymbol{\varepsilon}^e} : \dot{\boldsymbol{\varepsilon}}^e = \boldsymbol{\sigma} : \dot{\boldsymbol{\varepsilon}}^e.
\end{equation}
Substituting into the Clausius--Duhem inequality:
\begin{equation}
\boldsymbol{\sigma} : (\dot{\boldsymbol{\varepsilon}}^e + \dot{\boldsymbol{\varepsilon}}^p) - \boldsymbol{\sigma} : \dot{\boldsymbol{\varepsilon}}^e \ge 0.
\end{equation}
The elastic terms cancel, leading to the reduced inequality:
\begin{equation}
\mathcal{D} = \boldsymbol{\sigma} : \dot{\boldsymbol{\varepsilon}}^p \ge 0.
\end{equation}
This is the \textit{plastic dissipation inequality}, which states that the portion of mechanical power not stored elastically must be dissipated as irreversible plastic work.

\subsection*{B.2 Plastic Flow Rule for the KCC Model}

For the present derivation, we consider the plastic flow rule of the KCC model in which the plastic potential is written as
\begin{equation}
\phi(\boldsymbol{\sigma}) = \sqrt{3J_2(\mathbf{s})} - \omega\,\Gamma(p, J_3, \lambda),
\end{equation}
where $J_2$ is the second invariant of the deviatoric stress $\mathbf{s}$, $\omega$ is the non-associativity parameter, and $\Gamma$ represents the failure surface function depending on the mean stress $p$, the third invariant $J_3$, and an internal damage variable $\lambda$. The plastic strain rate is determined by
\begin{equation}
\dot{\boldsymbol{\varepsilon}}^p = \dot{\mu} \frac{\partial \phi}{\partial \boldsymbol{\sigma}}, \qquad \dot{\mu} \ge 0,
\end{equation}
where $\dot{\mu}$ is the plastic consistency parameter.
Applying the chain rule to the potential function gives
\begin{equation}
\frac{\partial \phi}{\partial \boldsymbol{\sigma}} 
= \frac{\partial}{\partial \boldsymbol{\sigma}}\sqrt{3J_2}
- \omega \left( 
\Gamma_{,p}\frac{\partial p}{\partial \boldsymbol{\sigma}}
+ \Gamma_{,J_3}\frac{\partial J_3}{\partial \boldsymbol{\sigma}}
+ \Gamma_{,\lambda}\frac{\partial \lambda}{\partial \boldsymbol{\sigma}}
\right),
\end{equation}
where the derivatives of $J_2$ and $p$ are
\begin{equation}
\frac{\partial}{\partial\boldsymbol{\sigma}}\sqrt{3J_2} = \frac{3}{2}\frac{\mathbf{s}}{\sqrt{3J_2}}, 
\qquad
\frac{\partial p}{\partial \boldsymbol{\sigma}} = -\frac{1}{3}\mathbf{I}.
\end{equation}
Neglecting the dependencies on $J_3$ and $\lambda$, the gradient of the potential simplifies to
\begin{equation}
\frac{\partial \phi}{\partial \boldsymbol{\sigma}} 
= \frac{3}{2}\frac{\mathbf{s}}{\sqrt{3J_2}} + \frac{\omega \Gamma_{,p}}{3}\mathbf{I}.
\end{equation}
The corresponding plastic flow rule is thus written as
\begin{equation}
\dot{\boldsymbol{\varepsilon}}^p = \dot{\mu}\left(
\frac{3}{2}\frac{\mathbf{s}}{\sqrt{3J_2}} + \frac{\omega \Gamma_{,p}}{3}\mathbf{I}
\right),
\end{equation}
and in incremental form,
\begin{equation}
\Delta \boldsymbol{\varepsilon}^p =
\left( \frac{3}{2}\frac{\mathbf{s}}{\sqrt{3J_2}} + \frac{\omega \Gamma_{,p}}{3}\mathbf{I}\right)\Delta \mu.
\end{equation}
This equation is identical to the given plastic strain increment equation (Eq. (69) in \cite{wu2015numerical}).

\subsection*{B.3 Thermodynamic Constraint for the KCC Model}

Substituting the flow rule into the plastic dissipation inequality yields
\begin{equation}
\mathcal{D} = \boldsymbol{\sigma} : \dot{\boldsymbol{\varepsilon}}^p 
= \dot{\mu} \, \boldsymbol{\sigma} : 
\left( \frac{3}{2}\frac{\mathbf{s}}{\sqrt{3J_2}} + \frac{\omega \Gamma_{,p}}{3}\mathbf{I} \right).
\end{equation}
With $\boldsymbol{\sigma} = \mathbf{s} - p\mathbf{I}$, and noting that $\mathbf{s}:\mathbf{I}=0$ and $\mathbf{I}:\mathbf{s}=0$, we obtain
\begin{align}
\boldsymbol{\sigma}:\frac{\partial \phi}{\partial \boldsymbol{\sigma}}
&= \frac{3}{2}\frac{\mathbf{s}:\mathbf{s}}{\sqrt{3J_2}} - p\omega\Gamma_{,p} \\
&= \sqrt{3J_2} - p\omega\Gamma_{,p}.
\end{align}
Thus, the plastic dissipation rate becomes
\begin{equation}
\mathcal{D} = \dot{\mu}\left(\sqrt{3J_2} - \omega \Gamma_{,p} p\right) \ge 0.
\end{equation}
This inequality expresses the thermodynamic requirement that the total rate of plastic work must be non-negative. It leads directly to the condition
\begin{equation}
\sqrt{3J_2} \ge \omega \Gamma_{,p} p,
\end{equation}


\section{Complete $\Gamma$–$\varepsilon_s$ Relations}
\label{appendix:complete set}
This appendix presents the full set of $\Gamma$–$\varepsilon_s$ curves for all confinement pressures ranging from $P_c = 5$ to \SI{39}{MPa} across all training conditions presented in Section~\ref{sec:demonstration}.
\begin{figure}[!ht]
\centering
\includegraphics[width=1\linewidth]{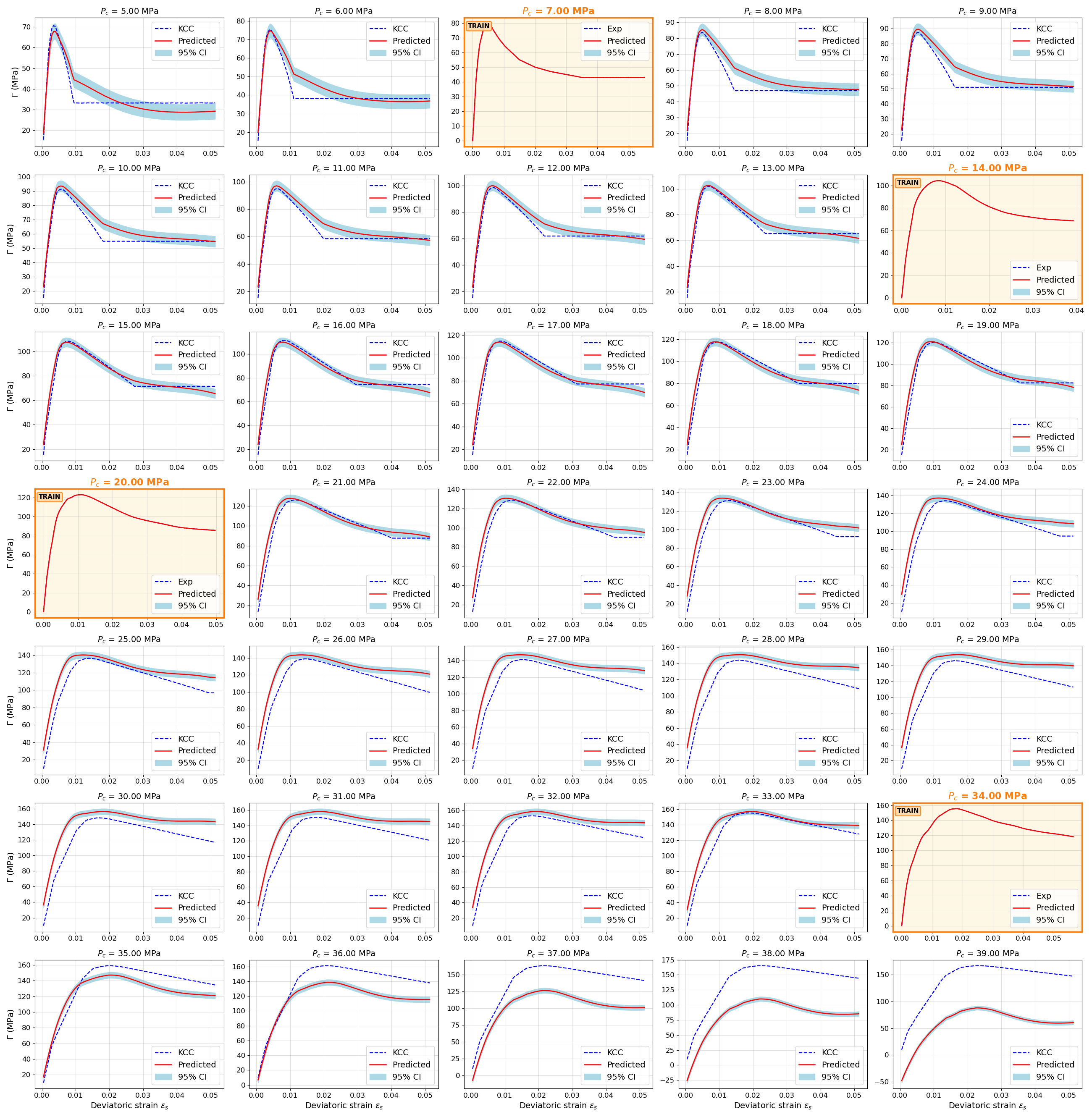}
\caption{Unconstrained GPR constitutive model trained using four experimental datasets showing $\Gamma$–$\varepsilon_s$ relations across a wide range of confinement levels from 5 to \SI{39}{MPa}. Experimental training data are highlighted in orange.}
\label{Train4_Reg}
\end{figure}

\begin{figure}[!ht]
\centering
\includegraphics[width=1\linewidth]{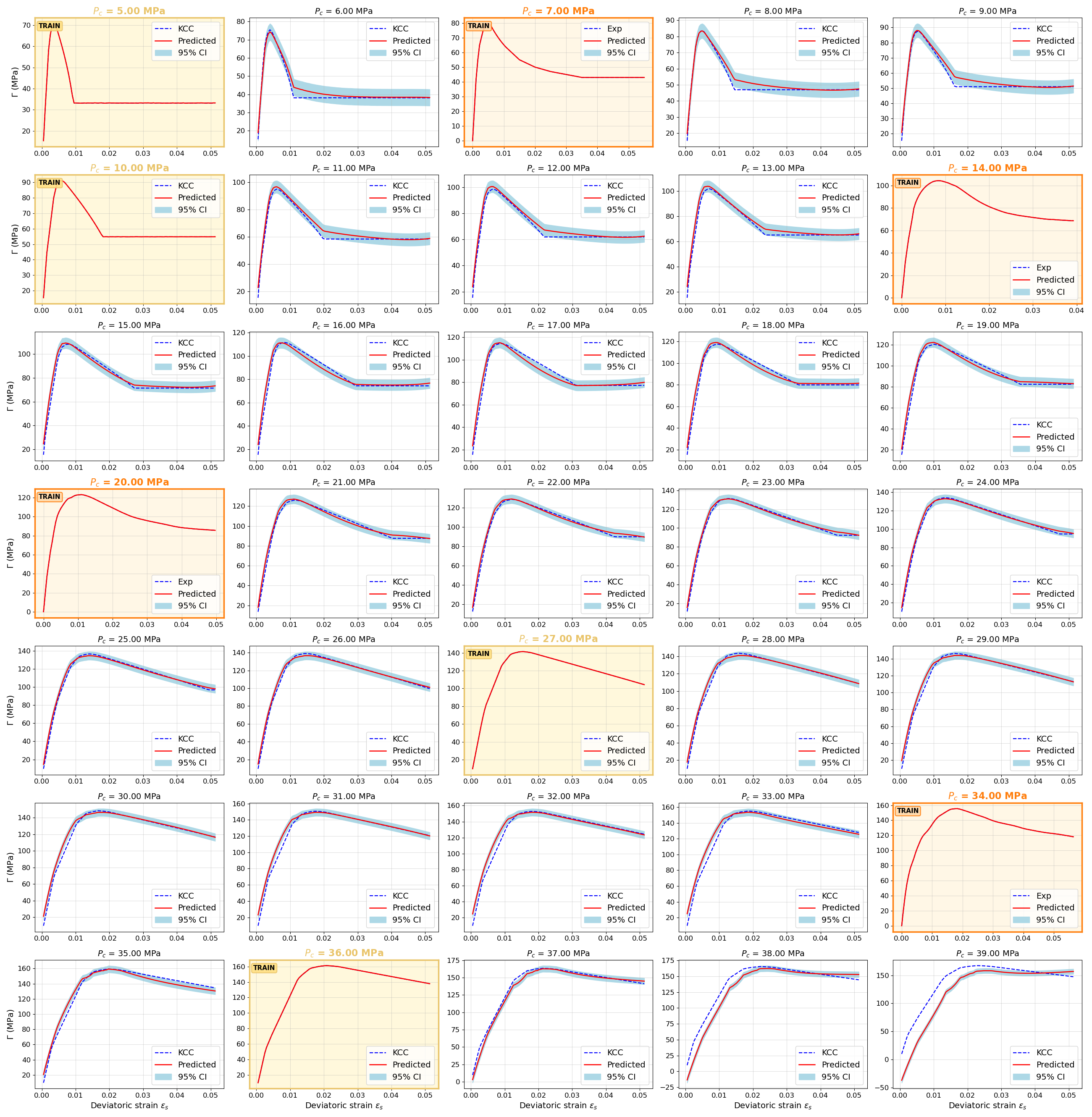}
\caption{Unconstrained GPR constitutive model trained using four experimental and four simulation datasets showing $\Gamma$–$\varepsilon_s$ relations across a wide range of confinement levels from 5 to \SI{39}{MPa}. Experimental raining data are highlighted in orange and simulated training datasets are highlighted in yellow.}
\label{Train8_Reg}
\end{figure}

\begin{figure}[!ht]
\centering
\includegraphics[width=1\linewidth]{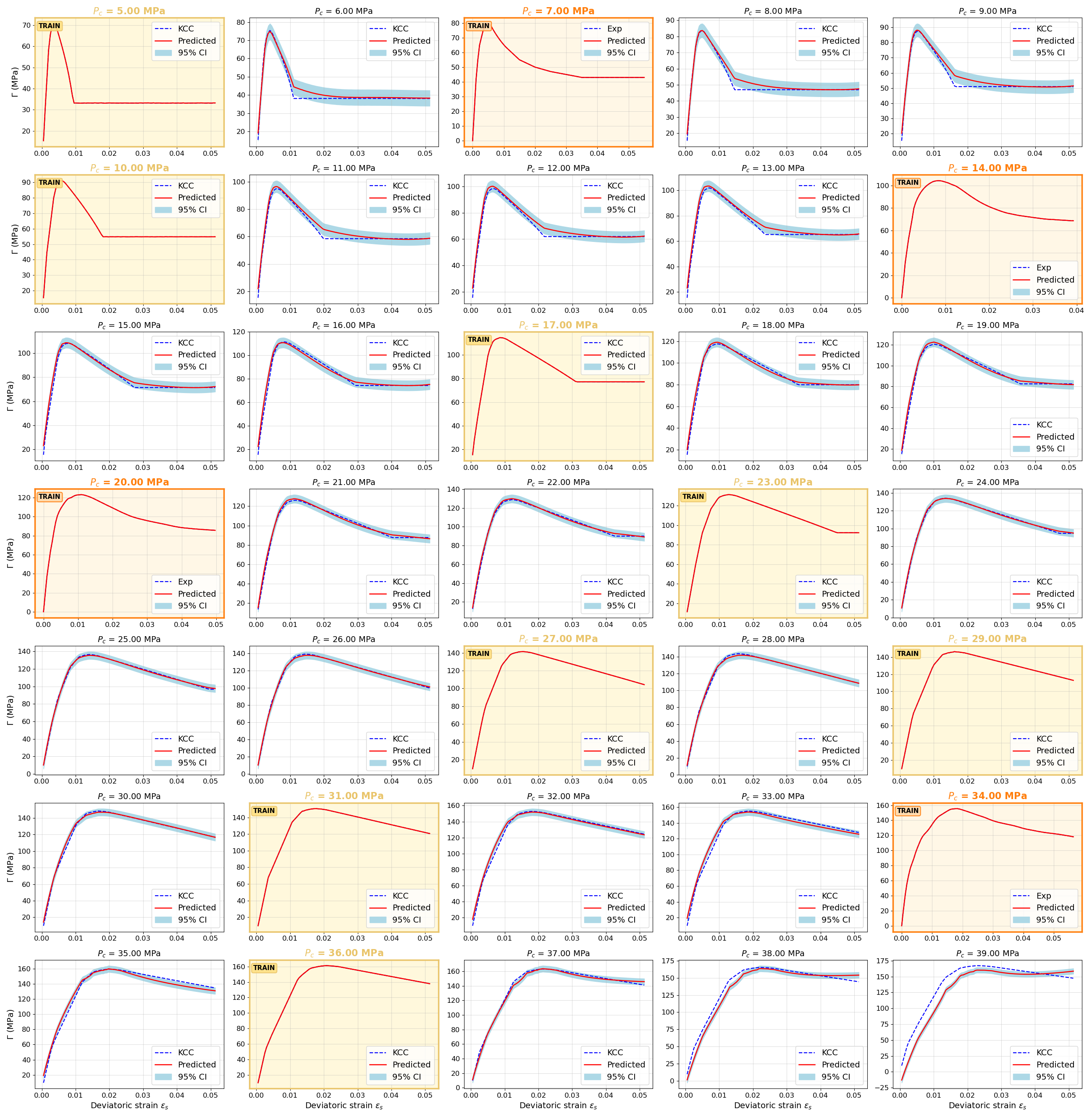}
\caption{Unconstrained GPR constitutive model trained using four experimental and eight simulation datasets showing $\Gamma$–$\varepsilon_s$ relations across a wide range of confinement levels from 5 to \SI{39}{MPa}. Experimental training data are highlighted in orange and simulated training datasets are highlighted in yellow.}
\label{Train12_Reg}
\end{figure}

\begin{figure}[!ht]
\centering
\includegraphics[width=1\linewidth]{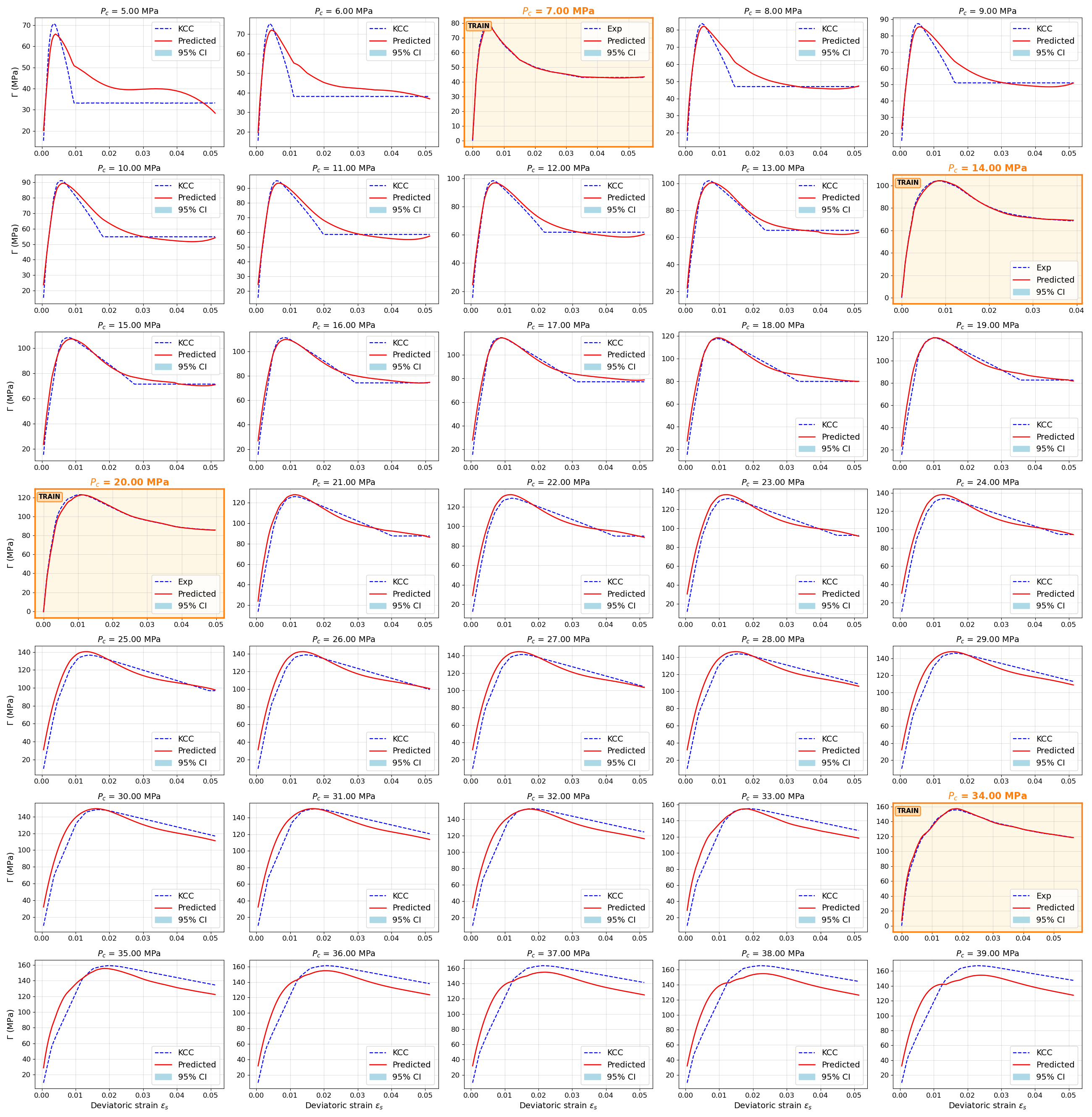}
\caption{Constrained GPR constitutive model trained using four experimental datasets showing $\Gamma$–$\varepsilon_s$ relations across a wide range of confinement levels from 5 to \SI{39}{MPa}. Experimental training data are highlighted in orange.}
\label{Train4_Con_Reg}
\end{figure}

\end{document}